\documentclass[a4paper,fleqn]{cas-dc}

\usepackage[authoryear]{natbib}

\usepackage{amsmath,amsfonts}
\usepackage{algorithm}
\usepackage{algpseudocode}
\usepackage{array}
\usepackage{minted}  
\usepackage{xcolor}

\newcommand{\revblue}[1]{#1}

\def\tsc#1{\csdef{#1}{\textsc{\lowercase{#1}}\xspace}}
\tsc{WGM}
\tsc{QE}

\begin{document}
\let\WriteBookmarks\relax
\def\floatpagepagefraction{1}
\def\textpagefraction{.001}

\shorttitle{RulerNet: Learning Perspective-Invariant Ruler Representations for Robust Image Scale Estimation}    

\shortauthors{Yimu Pan et al.}  

\title [mode = title]{RulerNet: Learning Perspective-Invariant Ruler Representations for Robust Image Scale Estimation}  

\tnotemark[1] 

\tnotetext[1]{Project Website: \url{https://github.com/ymp5078/RulerNet}.} 

%

\author[1]{Yimu Pan}[orcid=0009-0007-1025-1945]
\cormark[1]
\ead{ymp5078@psu.edu}
\author[1]{Manas Mehta}[orcid=0009-0003-2790-9601]
\author[2]{Gwen Sincerbeaux}
\author[3]{Jeffery A. Goldstein}[orcid=0000-0002-4086-057X]
\author[2]{Alison D. Gernand}[orcid=0000-0002-5661-7785]
\author[1]{James Z. Wang}[orcid=0000-0003-4379-4173]
\cormark[1]
\cortext[1]{Corresponding author}


\ead{jwang@ist.psu.edu}



\affiliation[1]{organization={Department of Informatics and Intelligent Systems, College of Information Sciences and Technology, The Pennsylvania State University},
            city={University Park},
            postcode={16802}, 
            state={PA},
            country={USA}}





\affiliation[2]{organization={Department of Nutritional Sciences, College of Health and Human Development, The Pennsylvania State University},
            city={University Park},
            postcode={16802}, 
            state={PA},
            country={USA}}

\affiliation[3]{organization={Department of Pathology, Feinberg School of Medicine, Northwestern University},
            city={Chicago},
            postcode={60611}, 
            state={IL},
            country={USA}}



\begin{abstract}
Accurately converting pixel measurements into absolute real-world dimensions remains a fundamental challenge in computer vision, limiting progress in applications such as biomedicine, forensics, nutritional analysis, and e-commerce. We introduce \textbf{RulerNet}, a deep learning framework that robustly infers scale in the wild by reformulating ruler reading as a unified keypoint detection problem and \revblue{representing rulers with geometric progression parameters that compactly approximate the non-uniform spacing induced by perspective transformations.} Unlike traditional methods that rely on handcrafted thresholds or rigid, ruler-specific pipelines, RulerNet directly localizes centimeter markings using a \revblue{mark-visibility-based annotation and training strategy that remains valid under mark-preserving transformations}, enabling strong generalization across diverse ruler types and imaging conditions while mitigating data scarcity. Additionally, we introduce a scalable synthetic data generation pipeline that combines graphics-based ruler creation with ControlNet-enhanced realism, significantly expanding training diversity and improving model performance.
Extensive experiments on both our datasets and public benchmarks demonstrate that RulerNet achieves accurate, consistent, and efficient scale estimation under challenging real-world conditions. Integration into a medical analysis pipeline further demonstrates its practical utility for scale-aware measurement. These results suggest that RulerNet can serve as a generalizable measurement component and be readily integrated with other vision modules, such as segmentation, monocular depth estimation, and diagnostic workflows, for automated analysis in medical and other domains. An online demo is provided. The code and data will be publicly released with the paper.
\end{abstract}




\maketitle

\section{Introduction}
\label{sec:intro}

\begin{figure}
    \centering
    \includegraphics[width=\linewidth]{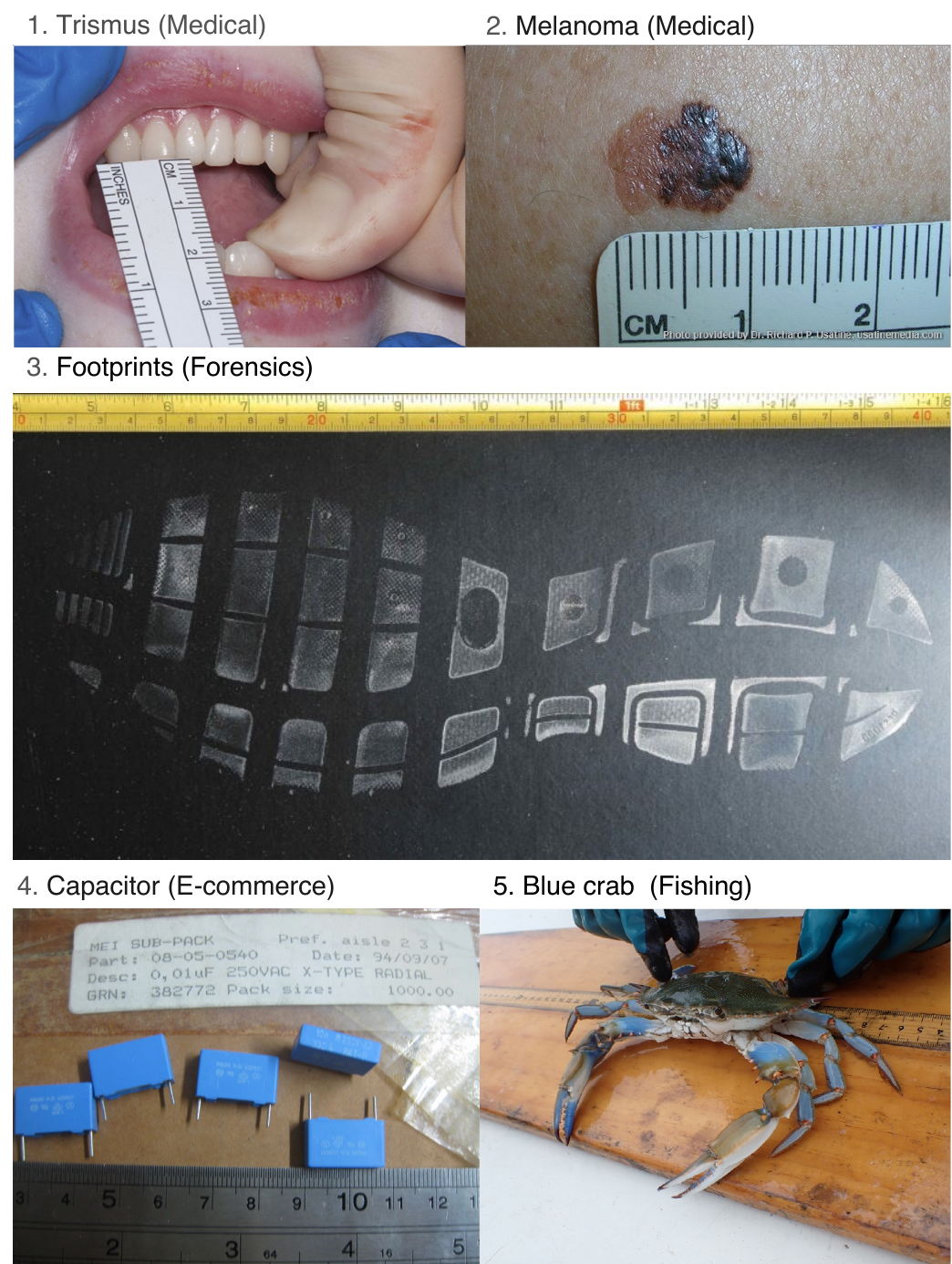}

    \caption{Representative real-world applications where accurate ruler-based scale estimation is essential.  1. Measuring mouth opening size aids in diagnosing Trismus, a condition that restricts jaw movement~\citep{Kresha_2019}. 2. Determining lesion size helps identify melanoma, where size progression is a critical diagnostic cue~\citep{American_Cancer_Society}. 3. Estimating human height from footprints is a common practice in forensic analysis and crime scene reconstruction~\citep{Crime_Museum_2023}. 4. Providing object size references in product images improves consumer trust and reduces returns in online shopping~\citep{eBay}. 5. Measuring fish length supports ecological surveys and fishery management by enabling accurate population estimates~\citep{Dep.nj.gov}. These examples highlight the broad societal relevance of accurate, automated measurement systems, motivating the need for robust, generalizable ruler reading solutions.}
    \label{fig:example_apps}
\end{figure}

Modern computer vision systems have achieved remarkable success in both low-level~\citep{minaee2021image} and high-level~\citep{grauman2022ego4d} recognition tasks. A fundamental challenge persists: converting pixel measurements into absolute, real-world dimensions. This capability, known as scale estimation, is critical for scientific research and industrial applications. Accurate scale estimation \revblue{bridges} the gap between relative spatial patterns and precise, quantitative measurements. For instance, in medical contexts, it supports disease diagnosis~\citep{abbasi2008utility} and informs surgical decision-making~\citep{thomas2003excision}; 
in e-commerce, especially within the fashion industry, it not only simplifies the size selection process but also personalizes recommendations, leading to a more seamless shopping experience~\citep{karessli2019sizenet,eshel2021presize,chatterjee2022incorporating};
and in health and nutrition research, it facilitates accurate food weight estimation from images~\citep{zhang2022eliminate,konstantakopoulos2023novel}. Although deep neural networks excel at learning relative spatial patterns, they inherently lack an intrinsic understanding of the metric scale, underscoring the pressing need for robust scale estimation techniques.

To overcome this limitation, practitioners often introduce reference objects~\citep{konstantakopoulos2023novel} into scenes as metric anchors, with rulers being the most universal and accessible choice. However, deriving accurate physical dimensions from photographs using ruler-based scale calibration remains a complex challenge. Although such calibration is common in biomedical imaging~\citep{chen2020ai,mirzaalian2019measuring}, urban planning~\citep{lin2018automatic,an2023water,yan2024soil}, and forensics~\citep{gertsovich2018automatic,wen2024ruler}, existing methods struggle to generalize to diverse real-world scenarios. The same challenge arises in our PlacentaVision\footnote{http://www.placentavision.com/} project, where determining the physical size of medical specimens is essential for downstream clinical analysis. Examples of other ruler-based measurement applications illustrating the complexity of imaging conditions and the diversity of rulers are given in Fig.~\ref{fig:example_apps}. Notably, the current state-of-the-art vision-language models (VLMs), such as GPT-4o~\citep{hurst2024gpt} with vision capabilities, are unable to correctly extract the ruler from Fig.~\ref{fig:example_apps}, highlighting the difficulty of this task even for advanced general-purpose models. Detailed qualitative results and failure examples are provided in the appendix. 

Traditional approaches~\citep{chen2020ai} rely on rigid, multi-step pipelines---ruler segmentation, intensity-based mark detection, and peak counting---that are prone to critical weaknesses:

\begin{itemize}
    \item \textit{Algorithmic fragility:} Reliance on handcrafted thresholds and heuristics to detect ruler marks renders these methods vulnerable to variations in illumination, contrast, and ruler appearance---such as differences in color, texture, or mark frequency---as well as geometric distortions like perspective warping.
    \item \textit{Inflexible scale estimation:} Estimating a single, uniform scale for an entire ruler overlooks perspective distortions. This simplification leads to significant errors, particularly when rulers are photographed at oblique angles.
    \item \textit{Dataset limitations:} Existing datasets typically include only a narrow range of ruler types, often standard metric rulers imaged under controlled conditions, due to the difficulty of collecting diverse ruler data. This leads to poor generalization to unconstrained, real-world scenarios involving unconventional rulers (e.g., minimalist printed or irregular scales, clear surfaces, glare or reflections, and white-on-black vs. black-on-white designs) and complex imaging conditions (e.g., shadows or occlusions).
\end{itemize}

\revblue{We propose \textit{RulerNet}, a unified deep learning framework that reimagines ruler reading as centimeter (cm) mark keypoint detection followed by compact geometric scale recovery. The central idea is that a straight physical ruler has equally spaced cm marks before imaging, while perspective projection makes those spacings non-uniform in the image. Rather than forcing projected marks to remain linearly spaced, RulerNet fits a geometric progression (GP) as a compact approximation to this perspective-induced spacing trend. This makes it possible to recover scale from unevenly spaced cm marks while avoiding brittle digit recognition, ruler-specific mark-frequency heuristics, or full camera calibration.}

\revblue{This GP-centered formulation distinguishes RulerNet from prior ruler-reading methods that depend on digit recognition, mark-frequency heuristics, or a single uniform image scale. RulerNet first maps detected two-dimensional cm marks onto a one-dimensional ruler coordinate and then estimates a compact GP representation of the projected mark sequence. The GP model is not treated as an exact projective invariant; rather, it provides an accurate low-dimensional approximation to the smooth spacing variation induced by perspective transformation under realistic camera-ruler geometry. This formulation preserves the straight-ruler constraint while allowing uneven projected mark spacing. DeepGP then learns the inverse mapping from noisy detected marks to the underlying GP structure, jointly estimating GP parameters, valid marks, adjacency, and point offsets so that missing, spurious, and locally shifted detections can be corrected during scale recovery.}

\revblue{Our \textbf{contributions} are:}
\begin{itemize}
    \item \revblue{A keypoint-based ruler-reading formulation that directly localizes cm marks and avoids hand-designed parsing of ruler appearance.}
    \item \revblue{A GP-based scale recovery model that represents perspective-induced non-uniform mark spacing with a small number of parameters after projecting detected marks to one dimension.}
    \item \revblue{DeepGP, a lightweight feed-forward module that learns to recover clean GP structure from noisy, missing, or spurious mark detections, enabling efficient inference.}
    \item \revblue{A hybrid synthetic-data pipeline that generates graphics-based rulers online during training and uses pre-generated ControlNet-refined generative rulers to improve coverage of ruler styles, imaging conditions, and mark-preserving transformations with limited manual annotation.}
    \item \revblue{Validation across diverse ruler datasets and a medical image-analysis pipeline, demonstrating that RulerNet can serve as a modular scale-estimation component for downstream measurement tasks.}
\end{itemize}

\section{Related Work}
\subsection{Ruler Detection and Reading}
Object scale measurement methods~\citep{jun2015research,telahun2020heuristic,lukashchuk2022method,matuzevivcius2023rulers2023} typically assume a standardized ruler design---uniform numbering, consistent millimeter (mm) marks, and other predictable features---even though they are not tailored to specific applications. In practice, many rulers deviate from these assumptions. Existing approaches often rely on handcrafted features, thresholding techniques based on ruler mark frequencies, or object detection methods designed to identify specific patterns (e.g., digits) tailored to certain ruler types. These methods have been applied in domains including fish studies~\citep{konovalov2017ruler}, forensic analysis~\citep{bhalerao2014ruler,gertsovich2018automatic,wen2024ruler}, museum cataloging~\citep{herrmann2010image}, urban flood monitoring and water level estimation~\citep{lin2018automatic,zhang2019situ,chen2021method,bai2021intelligent,dou2022research,qiu2023two,liu2023novel,an2023water}, medical imaging~\citep{goossen2008ruler,chen2015ruler,mirzaalian2019measuring,chen2020ai,xue2022extraction,xue2022automatic}, and soil profiling~\citep{yan2024soil}. In contrast, our method is designed to accurately read diverse rulers encountered in real-world settings.

\subsection{Data Generation and Synthesis}
Data generation and synthesis techniques are widely used to increase dataset variation and improve model performance and generalization. From an application perspective, these methods are closely related to data augmentation. Consequently, we discuss both data augmentation and generation in this section. Given the extensive literature on the topic, we focus on the aspects most relevant to our application and refer readers to a comprehensive survey~\citep{bauer2024comprehensive} for details.

Computer graphics-based techniques typically involve manipulating base shapes. Some approaches are implemented within visual environments~\citep{ros2016synthia,gaidon2016virtual,richter2016playing}, while others generate data directly~\citep{matuzevivcius2007mathematical,peris2012towards,rematas2014image,papon2015semantic,li2023paralleleye} to enhance training data diversity.
Generative approaches such as generative adversarial networks (GANs)~\citep{antoniou2017data,zhu2017data,bowles2018gan,shin2018medical,zhu2018emotion,bailo2019red,sandfort2019data,motamed2021data} have been used to create more realistic images. More recently, diffusion models~\citep{trabucco2023effective,feng2023diverse,zhang2023diffusion,chen2024synthetic,fang2024data,fu2024dreamda,islam2024diffusemix,alimisis2024advances} have been widely utilized for generating high-fidelity, detailed images~\citep{alimisis2024advances}.

Given the scarcity of ruler data, it is essential to incorporate both augmentation and data generation techniques. However, existing synthetic ruler approaches~\citep{matuzevivcius2023rulers2023} have relied solely on computer graphics-based methods. In this work, we extend ruler synthesis by incorporating generative methods, bridging the gap between the two approaches.

\section{Methods}
\begin{figure*}[!t]
    \centering
    \includegraphics[width=\linewidth]{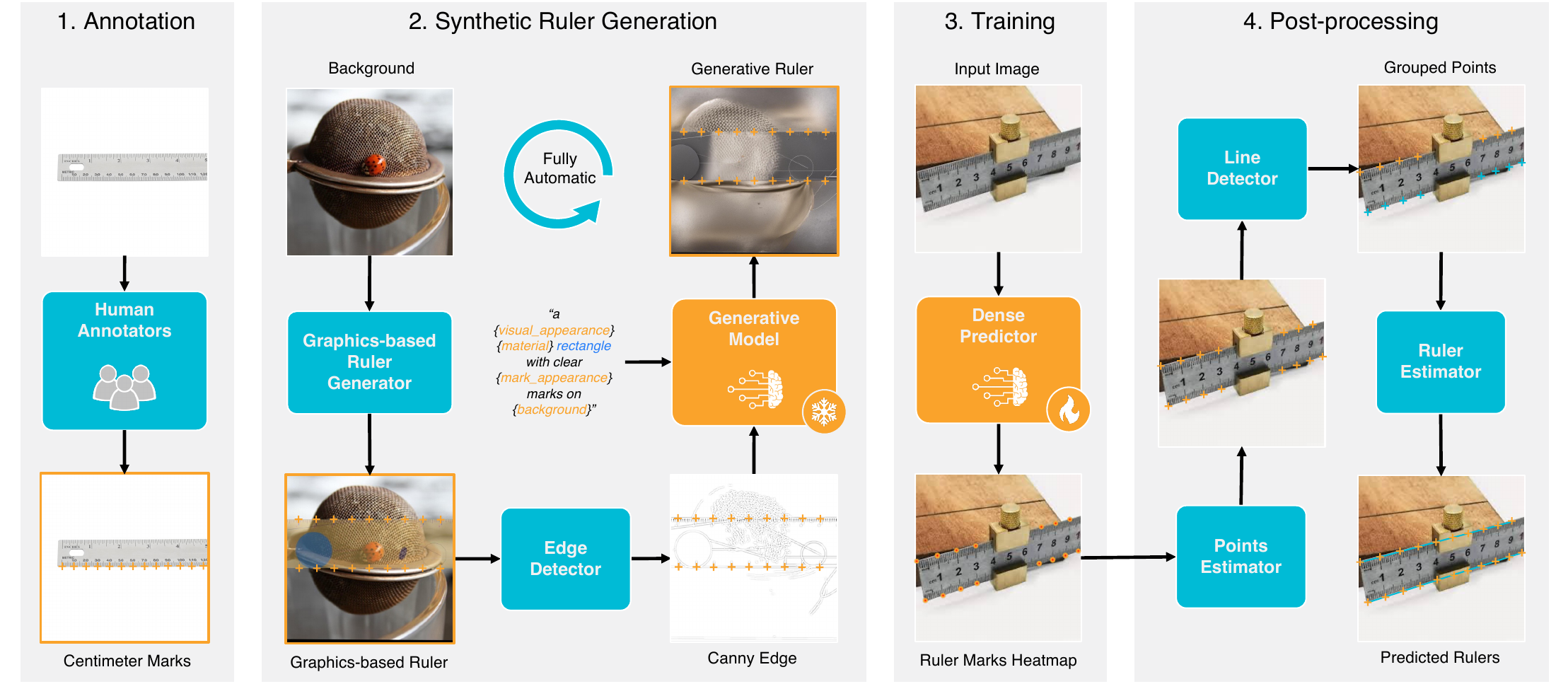}
    \caption{Overview of the RulerNet pipeline. Annotators label cm marks (orange ``+'') on rulers. A graphics-based generator synthesizes rulers on given backgrounds using specified parameters, followed by refinement with a frozen pre-trained generative model using its Canny edge map. A dense predictor is trained on both annotated and synthetic rulers to detect the marks, with post-processing applied to extract the final ruler (orange ``+'' with blue lines). Training data for the dense predictor is highlighted with an orange glow.}
    \label{fig:main}
\end{figure*}
The high-level development pipeline for RulerNet is shown in Fig.~\ref{fig:main}. Each stage is described in detail in the following subsections.
\subsection{Ruler Detection and Reading Reformulation}
Assuming minimal lens distortion (reasonable in practice, as per-lens calibration can be performed) and minimal perspective distortion (which is not reasonable, since the camera cannot be constrained to be parallel to the ruler), existing ruler-reading methods either target specific ruler designs or rely on frequency-based approaches. In a typical frequency-based pipeline, a ruler localization module (e.g., a segmentation model) first isolates the ruler within the image, after which a handcrafted module analyzes the frequency of marks within the extracted region. Although deep learning–based segmentation benefits from large training data, the subsequent handcrafted mark-reading step—which often assumes specific ruler appearances—remains a bottleneck for generalization.

Rather than following the traditional pipeline that addresses localization and reading separately, we remove the minimal perspective-distortion assumption and reformulate the problem as a keypoint detection task that simultaneously localizes and interprets ruler marks. In practice, cm marks are \revblue{widely} used and are usually visible even when mm marks vary in frequency or are partially obscured. Moreover, local regions around cm marks exhibit distinct visual characteristics compared to mm marks, which are usually shorter or thinner, or inch marks, which follow a different spatial frequency. Based on these observations, by directly predicting the positions of cm marks and ignoring the rest, we eliminate the need for a separate ruler localization step and reduce dependence on ruler-specific assumptions. \revblue{Accordingly, the current RulerNet model is trained for metric rulers, treating cm marks as target keypoints while treating mm and inch marks as non-target marks to be ignored.}

We implemented this approach with a heatmap-based model~\citep{wang2020deep} to predict the marks, allowing us to integrate an off-the-shelf dense prediction model. \revblue{Because detector training is formulated as a heatmap prediction for visible ruler-mark keypoints, the supervision itself does not require a rigid or straight ruler model. Any mark-preserving image transformation can be used provided that the target marks remain visible and their coordinates are transformed consistently. This requirement for mark visibility greatly expands the pool of usable real and synthetic training images, helping alleviate ruler-data scarcity.} Although our method is tailored for metric rulers with cm markings---owing to their wide adoption---the same data collection and training procedure can be readily adapted to imperial (inch) rulers if needed.

\subsection{Dataset Creation}
The quantity and quality of data play a crucial role in model performance. Although natural images are abundant, ruler images are relatively rare. To address this, we constructed a dataset that combines real and synthetic images. Additionally, we introduced a novel annotation scheme inspired by keypoint detection, which facilitates seamless integration into in-the-wild ruler-reading models.

\noindent\textit{Rulers from the Web:}
We used a scripted web search to collect metric ruler images and discarded those lacking a ruler, containing only inch markings, or featuring unreadable rulers. Because our annotation process and model training had to accommodate a wide variety of ruler designs, we encountered challenges such as inconsistent mark frequencies, blurred marks, low-resolution images that obscured mm marks, and perspective distortions affecting local scale. To address these issues, we directly annotated the positions of ruler marks as keypoints; the ruler scale is then computed as the distance between adjacent marks. The annotation procedure is illustrated in the first column of Fig.~\ref{fig:main}, with additional samples provided in the Appendix. 

\noindent\textit{Graphics-based Synthetic Rulers:}
We developed an algorithm that generates synthetic rulers superimposed on random images. Although other graphics-based methods~\citep{matuzevivcius2023rulers2023} for generating ruler images exist, they are typically not open-source and fail to account for variations in ruler mark frequency or provide the ground-truth ruler mark locations required for our use case.

Our objective was to create a ruler generator that facilitates training a cm mark detector, independent of the specific appearance of the ruler marks. To achieve this, we defined a set of controllable parameters that capture the range of variations observed in real-world rulers. These parameters cover multiple aspects of the generated image: for the background, we adjust the image source, perspective distortions, and random shapes; for the ruler rectangle, we control attributes such as edge color, fill color, transparency, size, edge thickness, and location; for the ruler marks, we set parameters including size, color, frequency, type (cm or inch), text font, and text placement. \revblue{We used this parameterized generator to create labeled graphics-based ruler images online during training, with backgrounds drawn from ImageNet~\citep{deng2009imagenet} to enhance diversity. Further details are provided in the Appendix.}

\revblue{This generator serves two main purposes: (1) it creates graphics-based synthetic ruler samples with known mark annotations on the fly during training, and (2) it provides the structured ruler layouts used as the foundation for the ControlNet-refined synthetic ruler generation pipeline described below.}

\noindent\textit{Generative Synthetic Rulers:}
Despite advances in graphics-based methods, there remains a significant domain gap between synthetic rulers and real-world examples. The predefined parameters used in graphics-based generation limit the variations in aspects such as illumination, material, and texture---factors that are naturally present in real images. Recently, conditional generative models like ControlNet~\citep{zhang2023adding} have shown remarkable success in producing realistic images and have been used for synthetic data generation to boost model performance.

Adopting a similar strategy, we utilized a pre-trained ControlNet guided by Canny edge maps for more realistic ruler generation. Our approach involved three key steps:

\begin{itemize}
    \item Generate a synthetic ruler image using our graphics-based method.
    \item Extract the Canny edge map from the generated image.
    \item Create a descriptive prompt that captures the texture, material, and environmental details of the ruler image. The prompt, together with the Canny edge map, is used to generate a more realistic ruler image.
\end{itemize}

This hybrid approach leverages the strengths of both graphics-based generation and conditional generative modeling to bridge the domain gap and improve the realism of the synthetic data. The data-generation pipeline is illustrated in the second column of Fig.~\ref{fig:main}. Examples of the generated images are available in the Appendix.

\subsection{Methods}

\noindent\textit{Ruler Mark Localization:}
We formulated mark localization as a pixel-level binary classification task with probabilistic targets, represented as a heatmap. We adopted an HRNet-like~\citep{wang2020deep} approach, where each mark was modeled as a 2D Gaussian centered at its location. These Gaussians were then aggregated into a heatmap, where higher probability values indicated the likely presence of ruler marks. Additionally, generating these Gaussians helped smooth out annotation inaccuracies that arose from the difficulty of precisely clicking on the ruler marks during labeling. \revblue{This localization stage only requires the target cm marks to be sufficiently visible for annotation and detection; it does not require a straight or rigid ruler model.}

This formulation enables the use of any off-the-shelf dense prediction network as a keypoint detector. Formally, we train our model using a combination of cross-entropy (CE) and DICE losses, with the latter adapted to continuous targets by squaring the source and target in the denominator. The DICE loss is defined as:
\begin{equation}
    \mathcal{L_{\mathrm{DICE}}}(\mathbf{X},\mathbf{Y})=1-\frac{\sum_{i}x_iy_i}{\sum_{i}x_i^2+\sum_{i}y_i^2}\;,
\end{equation}
where $x_i$ and $y_i$ denote the $i$th elements of $\mathbf{X} and \mathbf{Y}$, respectively. The total loss is given by:
\begin{equation}\label{eqn:binary_mask_loss}
    \mathcal{L}(\mathbf{X},\mathbf{Y})=\lambda_{\mathrm{CE}}\mathcal{L}_{\mathrm{CE}}(\mathbf{X},\mathbf{Y}) + \lambda_{\mathrm{DICE}}\mathcal{L_{\mathrm{DICE}}}(\mathbf{X},\mathbf{Y})\;,
\end{equation}
where $\lambda_{\mathrm{CE}}$ and $\lambda_{\mathrm{DICE}}$ are the hyperparameters.

\revblue{We use CE+DICE because cm marks occupy only a very small fraction of the heatmap. In a controlled loss ablation, a plain, unweighted L1 heatmap-regression model was fully trained but collapsed to all-background predictions during evaluation, yielding no usable mark detections. We therefore did not treat unweighted L1/L2 regression losses as competitive alternatives under the same training settings. This result supports the use of a segmentation-style objective that explicitly addresses sparse foreground structures.}

During inference, the predicted heatmap is converted back into a point representation for scale estimation. Since the target heatmap is generated by placing 2D Gaussians around the ground-truth points, and assuming that the Gaussian parameters are chosen such that Gaussians from distinct points do not overlap (i.e., the variance is sufficiently small), mark positions can be accurately recovered by extracting the connected components or local maxima from the heatmap.

\noindent\textit{Hough Transform for Ruler Detection:}
To accurately recover the scale during inference, two issues must be addressed. First, an image may contain multiple rulers. Second, some points might be misclassified as ruler marks. \revblue{At this grouping stage, we assume that the ruler used for scale estimation is locally straight, so that marks from the same physical ruler lie approximately along a common line in the image.} To group the detected ruler-mark points and filter out erroneous detections, we applied a Hough transform to the detected points, grouping them based on their linear alignment. Although other RANSAC-based linear regression methods can be used, they are more difficult to parallelize and require more detailed hyperparameter tuning to ensure a linear relationship for each group of points.

\noindent\textit{Scale Estimation:}
{
Unlike prior methods~\citep{lukashchuk2022method} that assume ruler marks follow a linear progression in the image, our approach generalizes to in-the-wild rulers by making the assumptions underlying scale recovery explicit. Specifically, we assume that (1) the physical ruler has equally spaced cm marks before imaging, (2) the detected marks have already been grouped along a locally straight ruler line, and (3) strong lens distortion is absent or has been corrected. Let $x_i$ denote the one-dimensional physical coordinate of the $i$th cm mark along the ruler, distinct from the image-space mark locations $m_i$ introduced below. The physical ruler marks satisfy
\begin{equation}
    x_i = x_0 + i\ell\;,
\end{equation}
where $\ell$ is the fixed physical distance between adjacent cm marks. Under perspective projection, these collinear physical marks remain collinear in the image, but their Euclidean image-space spacings are generally no longer uniform. Therefore, the equal-spacing assumption applies to the physical ruler coordinates rather than the projected image coordinates.

Instead of enforcing equal spacing in the image, we approximate the observed image-space ruler spacings using a GP. This relaxes the conventional linear-spacing assumption and allows the observed ruler marks to exhibit the monotonic compression or expansion induced by perspective projection. Although perspective projection preserves the collinearity of ruler marks, it generally does not preserve Euclidean distances or the ratios of adjacent distances. Therefore, the GP model should be interpreted as a compact approximation to the projective spacing patterns rather than as an exact projective invariant.

The approximation remains accurate under realistic camera geometries. Along a straight ruler, perspective projection reduces to a one-dimensional homography, $u_i=(a x_i+b)/(c x_i+d)$. If the physical cm spacing is $\ell$ and $\eta=c\ell/(c x_0+d)$ denotes the normalized perspective strength, the projected interval spacings, normalized by the first interval, satisfy
}
\begin{equation}
    D_i^{\mathrm{true}}
    =
    \frac{1+\eta}{(1+i\eta)(1+(i+1)\eta)}\;.
\end{equation}
{
A GP approximation models these normalized spacings as $D_i^{\mathrm{GP}}=\rho^i$. For $N$ visible intervals, choosing $\rho$ to match the first and last projected intervals yields
}
\begin{equation}
    \rho =
    \left(
    \frac{1+\eta}{(1+(N-1)\eta)(1+N\eta)}
    \right)^{\frac{1}{N-1}}\;.
\end{equation}
{
This yields a worst-case relative spacing error, defined as $E_{\max}=\max_i|D_i^{\mathrm{GP}}/D_i^{\mathrm{true}}-1|$. Under a pinhole camera model, $\eta=\ell\sin\phi/Z_0$, where $Z_0$ is the depth of the first visible mark and $\phi$ is the angle between the direction of the ruler and the image plane. For $\ell=1$ cm and $N=10$ visible intervals, even under the conservative strong-perspective limit $\lim_{\phi\to90^\circ}\sin\phi=1$, the worst-case spacing errors are only 2.30\%, 1.68\%, 1.01\%, and 0.67\% for $Z_0=25$, 30, 40, and 50 cm, respectively. For context, in our typical desktop imaging setting at a $768\times768$ input resolution, a ruler resolution of 30 pixels/cm means that a one-pixel measurement error corresponds to a relative scale error of $1/30=3.33\%$, which is larger than all of these GP approximation errors. These values support the use of GP as a simple approximation to projective ruler spacing in typical close-range imaging. A complete derivation is provided in the Appendix.

Given three consecutive cm mark points, $m_i$, $m_{i+1}$, and $m_{i+2}$, the GP approximation can be represented as:
}\begin{equation}\label{eqn:gp}
    \mathcal{D}_\mathrm{Euclidean}(m_{i+2},m_{i+1}) = r \mathcal{D}_\mathrm{Euclidean}(m_{i+1},m_{i})\;,
\end{equation}
{
where $\mathcal{D}_\mathrm{Euclidean}$ denotes the Euclidean distance, and $r$ is the common ratio. When $r=1$, the progression reduces to a uniform image-space spacing. When $r\neq 1$, the model captures the monotonic compression or expansion of the image-space ruler spacing using a single scalar degree of freedom. This provides a simple, yet effective, perspective-aware ruler representation while avoiding the strict assumption that ruler marks remain equally spaced in the image.
}

{
\revblue{To fully determine the ruler's GP representation, we estimate two adjacent recovered clean marks, $m_0$ and $m_1$, together with the common ratio $r$, from the noisy detected mark set $\mathbf{M}_{\text{noisy}}$. The optimization searches for the latent GP sequence that best explains the noisy detections, thereby recovering missing or displaced cm marks rather than treating the noisy detections as ground truth:}
}
\begin{equation}\label{eqn:opt}
    \min_{m_0, m_1, r} \quad \mathcal{D}_{\mathrm{Chamfer}}\left(\mathbf{M}_{\text{noisy}}, \mathbf{M}(m_0, m_1, r)\right)\;,
\end{equation}
{\revblue{subject to the constraints $r_{\min} < r < r_{\max}$ and 
$\mathcal{D}_{\min} < \mathcal{D}_{\mathrm{Euclidean}}(m_{1},m_{0}) < \mathcal{D}_{\max}$\;,
where $r_{\min}$, $r_{\max}$, $\mathcal{D}_{\min}$, and $\mathcal{D}_{\max}$ are hyperparameters. Here, $\mathbf{M}(m_0, m_1, r)$ denotes the clean ruler-mark sequence generated by the GP model, and the Chamfer objective allows this recovered sequence to remain close to the noisy detections while correcting localization errors, missing marks, and spurious detections.}
}

In the optimization process, the Chamfer distance---whose computational complexity is $\mathcal{O}(dn^2)$, where $d$ is the point dimension and $n$ is the number of points---is computed repeatedly. Therefore, a more efficient computation of the Chamfer distance would greatly improve overall optimization efficiency. Although existing methods~\citep{bakshi2023near} reduce complexity from $\mathcal{O}(dn^2)$ to $\mathcal{O}(n\log n)$, we further propose a simple technique for our case to reduce the effective dimension $d$ to 1.

Unlike arbitrary 2D points, our detected points lie along a line (as determined by the Hough transform). By precomputing the line representation, we project all points onto this line, effectively reducing the problem to a one-dimensional space. This allows us to compute the Chamfer distance in $\mathcal{O}(n^2)$ or $\mathcal{O}(n\log n)$ time, depending on the algorithm used.

To obtain the parameters $m_0$, $m_1$, and $r$ that best represent the ruler, we employ differential evolution~\citep{storn1997differential}, a global optimization technique, as a baseline method for performing this optimization.

Using the GP representation of the ruler, we can compute the length of any line segment along the ruler without requiring an undistorted camera view, making our method significantly more generalizable than prior approaches. When marks are partially occluded, the model can still detect adjacent points to recover the ruler via GP extrapolation. This is advantageous in real-world scenarios, where rulers are rarely imaged in perfect alignment with the camera.

\noindent\textit{Robustness Improvement and Inference Acceleration:}  
Although the GP representation significantly improves generalization, the per-sample optimization strategy based on Chamfer distance presents two major limitations. First, it is sensitive to noise in detected ruler marks, including missing points and spurious detections. Second, the iterative optimization required at inference time is computationally expensive, limiting scalability and real-time deployment.

A natural alternative is to directly regress the GP parameters $(m_0, m_1, r)$ from noisy ruler mark detections. However, directly regressing the initial points $(m_0, m_1)$ is fundamentally ambiguous and problematic: multiple pairs of initial points may be equally valid as long as they correspond to adjacent elements of the underlying GP.

To address this issue, we reformulate GP estimation as a structured prediction problem. Instead of directly predicting initial points, the model predicts (i) a \emph{validity mask} indicating which detected points belong to the GP, (ii) an \emph{adjacency mask} identifying which neighboring points form consecutive GP elements, and (iii) per-point \emph{offsets} that correct localization noise. This formulation removes ambiguity by allowing the model to implicitly infer the GP origin through mask-based reasoning over all candidate points.

We introduce a learned inverse mapping, termed \textit{DeepGP}, implemented using a 1D U-Net architecture. Given projected 1D ruler marks $\mathbf{M}_{\text{noisy}}$, the encoder extracts global contextual features, followed by two task-specific branches: a lightweight regression head that predicts the common ratio $r$, and a decoder branch that predicts the validity mask $\mathbf{v}$, the adjacency mask $\mathbf{a}$, and the offsets $\Delta \mathbf{M}$. The refined point locations are obtained as $\hat{\mathbf{M}} = \mathbf{M}_{\text{noisy}} + \Delta \mathbf{M}$:
\begin{equation}
    r,\ \mathbf{v},\ \mathbf{a},\ \Delta \mathbf{M}
    = \mathrm{DeepGP}(\mathbf{M}_{\text{noisy}})\;.
\end{equation}

DeepGP is trained end-to-end using a multi-task objective:
\begin{equation}
\mathcal{L}
=
\lambda_r \mathcal{L}_{\text{ratio}}
+
\lambda_v \mathcal{L}_{\text{valid}}
+
\lambda_a \mathcal{L}_{\text{adj}}
+
\lambda_m \mathcal{L}_{\text{offset}}\;,
\end{equation}
where $\mathcal{L}_{\text{ratio}}$ is an $\ell_1$ loss on the predicted ratio, and $\mathcal{L}_{\text{valid}}$ and $\mathcal{L}_{\text{adj}}$ are defined in Equation~\ref{eqn:binary_mask_loss}. The offset loss $\mathcal{L}_{\text{offset}}$ supervises the localization only at valid points, preventing outliers from dominating training.

To enable robust training at scale, we generate synthetic data by randomly sampling GP parameters $(m_0, m_1, r)$ and constructing the corresponding ruler marks, as described in Algorithm~\ref{alg:noisy_gp}. Realistic noise is introduced through point perturbation, random point dropout, and injection of spurious detections. Since ruler marks lie on a straight line, all 2D detections are projected to 1D prior to inference, reducing computational cost while preserving geometric structure.

The initial GP points $(m_0, m_1)$ are recovered by selecting the most reliable pair of adjacent valid points. Specifically, for each neighboring pair $(i,i{+}1)$, we compute a confidence score
\begin{equation}
    s_i = a_i + v_i + v_{i+1}\;,
\end{equation}
and select $i^\ast = \arg\max_i s_i$. The initial points are then given by
\begin{equation}
    m_0 = \hat{m}_{i^\ast}, \quad m_1 = \hat{m}_{i^\ast+1}\;.
\end{equation}

By eliminating iterative optimization and resolving GP ambiguity through structured mask prediction, DeepGP enables fast, robust, and scalable geometric-progression parameter estimation. The formulation is architecture-agnostic and readily transferable to other 1D structure detection tasks involving repeated or progressively scaled patterns under noise and partial observability.


\begin{algorithm}[!ht]
\caption{Generate Noisy Geometric Progression Points with Structural Supervision}
\label{alg:noisy_gp}
\begin{algorithmic}[1]
\Require Initial points $m_0, m_1$, geometric ratio $r$, target length $N$
\Ensure $\mathbf{M}_{\text{noisy}}\in\mathbb{R}^{N}$, validity mask $\mathbf{v}\in\{0,1\}^{N}$, adjacency mask $\mathbf{a}\in\{0,1\}^{N-1}$, offsets $\Delta\mathbf{M}\in\mathbb{R}^{N}$

\State Initialize clean point list $M = [\,]$ and index list $I = [\,]$
\State Append $m_0,m_1$ to $M$ and append indices $0,1$ to $I$
\For{$k = 2$ to $N-1$}
    \State $m_k \gets r (m_{k-1} - m_{k-2}) + m_{k-1}$
    \State Append $m_k$ to $M$ and append index $k$ to $I$
\EndFor
\Comment{$(M,I)$ defines clean GP points with integer IDs}

\State Add Gaussian noise to points:
\State $\tilde{M} \gets \{ m + \epsilon \mid m \in M,\ \epsilon \sim \mathcal{N}(0,\sigma^2) \}$
\State $\tilde{I} \gets I$

\State Randomly remove a subset of points from $\tilde{M}$ and remove corresponding entries from $\tilde{I}$

\State Sample $S$ spurious points $\{u_j\}_{j=1}^{S}$, $u_j \sim \mathcal{U}(\ell,h)$
\State Append spurious points to $\tilde{M}$ and append index $-1$ for each spurious point to $\tilde{I}$

\State Sort $\tilde{M}$ in ascending order and permute $\tilde{I}$ accordingly

\If{$|\tilde{M}| \ge N$}
    \State $\mathbf{M}_{\text{noisy}} \gets \tilde{M}_{1:N}$ and $\mathbf{idx} \gets \tilde{I}_{1:N}$
\Else
    \State Pad $\tilde{M}$ with zeros to length $N$ and pad $\tilde{I}$ with $-1$ to length $N$
    \State $\mathbf{M}_{\text{noisy}} \gets \tilde{M}$ and $\mathbf{idx} \gets \tilde{I}$
\EndIf

\State Set validity mask: $v_i \gets \mathbb{I}[\mathrm{idx}_i \ge 0] \quad \forall i$
\State Set adjacency mask:
\[
a_i \gets \mathbb{I}\big[(\mathrm{idx}_{i+1}-\mathrm{idx}_i = 1)
\wedge (\mathrm{idx}_i \ge 0)
\wedge (\mathrm{idx}_{i+1} \ge 0)\big]
\]

\State Initialize clean aligned points $\mathbf{M}_{\text{clean}} \gets \mathbf{0}$
\For{$i = 1$ to $N$}
    \If{$\mathrm{idx}_i \ge 0$}
        \State $\mathbf{M}_{\text{clean},i} \gets$ clean GP point with index $\mathrm{idx}_i$
    \EndIf
\EndFor
\State Compute offsets: $\Delta\mathbf{M} \gets \mathbf{M}_{\text{clean}} - \mathbf{M}_{\text{noisy}}$

\State \Return $\mathbf{M}_{\text{noisy}},\ \mathbf{v},\ \mathbf{a},\ \Delta\mathbf{M}$
\end{algorithmic}
\end{algorithm}

\noindent\textit{\revblue{Application to Object Size Measurement:}}
\revblue{RulerNet estimates image-plane scale from the detected ruler marks. To transfer this scale to an object measurement, the object should lie approximately in the same physical plane as the ruler, or the 3D relationship between the ruler and the object must be known so that the scale can be corrected. The GP model captures scale variation along the ruler direction, allowing measurements aligned with that direction to use the corresponding local scale. However, the one-dimensional ruler representation does not model depth variations or tilt perpendicular to the ruler direction. If these assumptions are violated, the recovered ruler scale may remain accurate for the ruler itself, but object-size estimates may become inaccurate.}

\section{Results}

\subsection{Training and Testing Sets}

\begin{figure}
    \centering
    \includegraphics[width=\linewidth]{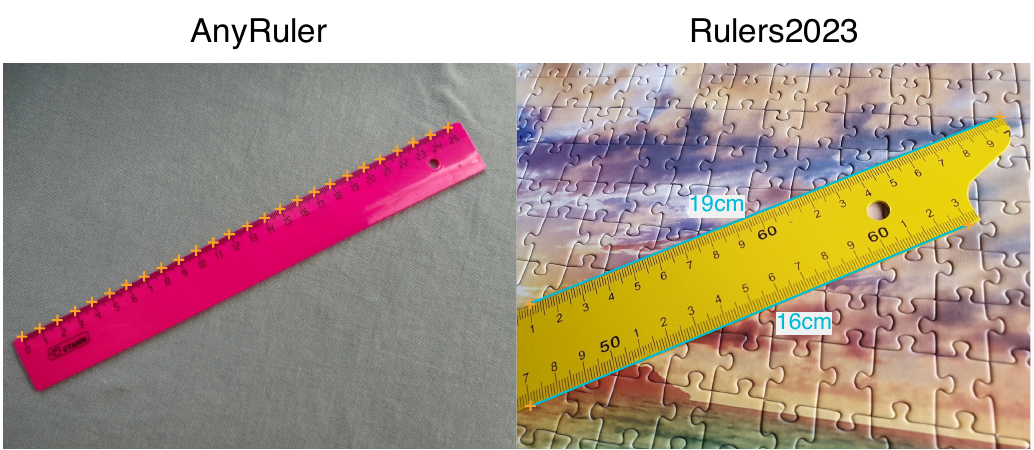}
    \caption{Example annotations from the AnyRuler and Rulers2023 datasets. Orange `+' markers indicate point annotations, while blue lines represent line annotations, with their corresponding lengths displayed alongside them.}
    \label{fig:annotations}
\end{figure}

We obtained 1,416 annotated ruler images with diverse appearances and image conditions from the internet, collectively referred to as the AnyRuler dataset. Approximately 80\% of these images were randomly selected for training, with the remaining 20\% reserved for testing. In addition to evaluating performance on a variety of rulers collected from the internet, we also aimed to assess RulerNet's generalizability to practical application scenarios. We used the Rulers2023 dataset~\citep{matuzevivcius2023rulers2023}, which contains ruler images captured under diverse conditions that differ significantly from those in the training dataset. Since this dataset was used solely for evaluation, we manually drew lines covering the cm marks and labeled the length of each line in the Rulers2023 test set, resulting in 654 images. Each sample was annotated by one annotator and verified by another. Example annotations are shown in Fig.~\ref{fig:annotations}.

For consistency in comparison, we extracted a pixels/cm value from each image. For the AnyRuler testing set, where the labels consisted of points on visible cm marks, we first computed the pixels/cm for all adjacent marks along the same ruler. We then took the median pixels/cm for each set of cm marks to reduce noise from missing marks. Finally, we assigned the pixels/cm value of the largest set of marks as the representative value for the image. For the Rulers2023 dataset, since we directly labeled the length of the line along the cm marks, we computed the pixels/cm for each ruler and used the value of the longest ruler for each image.

\subsection{Metric}
We measured the difference between the prediction and the ground truth scale. Mean Absolute Pixel Error per cm (mAPE/cm) was utilized as the metric, where lower values indicated better performance. Since mAPE/cm could be affected by the ground-truth pixels/cm, we scaled it according to the image resolution to ensure a fair comparison. Specifically, we defined $n$ as the base resolution and scaled the metric by the ratio of the current image resolution to $n$, termed mAPE/cm@$n$. Formally, it was defined as:
\begin{equation}
    \mathrm{mAPE/cm@}n = \frac{1}{|Q|}\sum_{i} \frac{n\|p_i-q_i\|}{s_i}\;,
\end{equation}
where $p_i \in \mathbf{P}$, $q_i \in \mathbf{Q}$, and $s_i \in \mathbf{S}$ represented the predicted scales, ground-truth scales, and image sizes for the $i$th image in the dataset, respectively. mAPE/cm@768 was used for all experiments unless specified otherwise. Since all results were reported on the same dataset across methods, the absolute cm scale did not affect the evaluation.

\revblue{If a method returned an incorrect, nonzero scale, we evaluated the predicted value directly. If a method failed to detect a valid ruler scale, we assigned its prediction a value of 0 pixels/cm for metric computation. This convention penalizes complete detection failure without modifying erroneous nonzero predictions, allowing both failure modes to contribute to mAPE/cm according to the resulting scale error.}

For inference speed evaluation, we report the average number of milliseconds per sample (ms/sample) across the entire dataset, excluding the first 10\% of samples to eliminate the effects of loading and caching overhead. To specifically isolate the computational cost of ruler extraction, we compare only the ruler extraction modules in the ablation study.

While our primary evaluation focuses on pixels/cm conversion accuracy, we acknowledge that this assumes co-planarity between the ruler and object. This evaluation choice reflects the current absence of publicly available datasets with full 3D annotations or camera intrinsics necessary for metric calibration through homography or depth estimation. Nevertheless, our AnyRuler dataset and Rulers2023 annotations—comprising a diverse range of ruler types, imaging conditions, and annotations—represent the most comprehensive public benchmark for ruler-based scale estimation to date. We believe that our work lays the foundation for future research that integrates monocular depth prediction, homography estimation, or 3D-aware measurement models.

\subsection{Implementation Details}

\noindent\textit{Training Setup:}  
{
\revblue{Our final RulerNet-DeepGP model used a MobileNetV4~\citep{qin2024mobilenetv4} encoder with a U-Net~\citep{ronneberger2015u} dense-prediction architecture at a resolution of $768 \times 768$, followed by DeepGP for ruler extraction. The dense mark detector used the same CE+DICE heatmap objective for both pre-training and training. Pre-training used the AnyRuler training set along with the synthetic rulers from both pipelines: the graphics-based rulers were generated on the fly by the programmatic generator, while the ControlNet-refined generative rulers were drawn from a fixed pre-generated dataset. We used 102,000 samples from each synthetic source, with the graphics-based samples generated on the fly and the generative samples drawn from the fixed pre-generated dataset. The subsequent training stage used the same convention, with the number and composition of graphics-based and generative synthetic samples varying across the ablation studies. The Canny detector used for generative image creation applied a low threshold of 100 and a high threshold of 200. The DeepGP module was implemented as a custom 1D U-Net with only 1.6M parameters, and was trained on 1.23 billion randomly generated 1D GP mark sequences using a batch size of 1,024 to ensure broad coverage of GP scenarios. We capped the number of ruler marks at 64, as higher counts (corresponding to rulers longer than 0.64 meters) result in cm intervals that are too small to be reliably distinguished. To standardize input for DeepGP, all mark locations were normalized using min-max normalization to the range \([-1, 1]\).}
}

Since no open-source implementations were available for existing ruler-reading methods, we selected two representative baselines for comparison: a digit detection-based method~\citep{lukashchuk2022method} and a mark frequency-based method~\citep{wen2024ruler}. The digit detection-based method was implemented using an off-the-shelf OCR model (EasyOCR\footnote{\url{https://github.com/JaidedAI/EasyOCR}}), and we incorporated localization and post-processing of the detected digit positions using our GP method to enable fair comparison. For the mark frequency-based method~\citep{wen2024ruler}, we obtained performance results directly from the original authors; however, inference time was not available.

All experiments were conducted using Python 3.9, PyTorch 2.0.1, scikit-learn 1.5.2, OpenCV 4.8.1, and NumPy 1.26.4. All RulerNet training was performed using consistent training settings on a single NVIDIA A40 GPU, except for pre-training, which was conducted for 80 epochs on four NVIDIA A40 GPUs at a base learning rate of 0.0005. All other models were trained for 200 epochs using the AdamW optimizer with default PyTorch settings and a cosine annealing learning rate scheduler, including a 10\% linear warmup phase at a base learning rate of 0.0002. We set $\lambda_{CE}=0.2$ and $\lambda_{DICE}=0.8$ for all experiments. 

\noindent\textit{Inference Setup:}  
All inference-time evaluations were performed on a cloud server equipped with an NVIDIA V100 GPU and an Intel Xeon Gold 6248 CPU.

\begin{algorithm}[ht!]
\caption{Extract Peak Points from Heatmap (PyTorch)}
\label{alg:extract_points_onnx}
\begin{algorithmic}[1]
\Require Heatmap $H \in \mathbb{R}^{H \times W}$, threshold $\tau=0.5$, kernel size $k$, Gaussian $\sigma$
\Ensure Coordinates $C$ of peak points

\Comment{Apply smoothing}
\State $K \gets$ predefined Gaussian kernel with $\sigma$
\State $S \gets \texttt{conv2d}(H, K, \text{padding} = k // 2)$

\Comment{Compute X-direction local maxima}
\State $dx \gets \texttt{diff}(S, \text{dim} = 1)$
\State $x_{\text{left}} \gets (dx[:, :-1] > 0)$
\State $x_{\text{right}} \gets (dx[:, 1:] < 0)$
\State $x_{\text{peaks}} \gets \texttt{pad}(x_{\text{left}} \land x_{\text{right}}, \text{pad} = (1, 1))$

\Comment{Compute Y-direction local maxima}
\State $dy \gets \texttt{diff}(S, \text{dim} = 0)$
\State $y_{\text{top}} \gets (dy[:-1, :] > 0)$
\State $y_{\text{bottom}} \gets (dy[1:, :] < 0)$
\State $y_{\text{peaks}} \gets \texttt{pad}(y_{\text{top}} \land y_{\text{bottom}}, \text{pad} = (1, 1))$

\Comment{Find valid peaks above threshold}
\State $P \gets x_{\text{peaks}} \land y_{\text{peaks}} \land (S > \tau)$
\State $C \gets \texttt{nonzero}(P)$

\State \Return $C$
\end{algorithmic}
\end{algorithm}

\begin{algorithm}[ht!]
\caption{Hough Line Detection (PyTorch)}
\label{alg:hough_line_onnx}
\begin{algorithmic}[1]
\Require Point set $P \in \mathbb{R}^{N \times 2}$, angle step $\Delta\theta=0.5$, resolution $\Delta\rho=5$
\Ensure Points $P_{\text{max}}$ supporting the dominant line

\Comment{Generate angle and rho candidates}
\State $\Theta \gets \texttt{deg2rad}(\texttt{arange}(-90, 90, \Delta\theta))$
\State $X, Y \gets P[:,0], P[:,1]$
\State $\rho \gets X \cos(\Theta) + Y \sin(\Theta)$ \Comment{Compute all $\rho_{i,j}$}

\State $\rho_{\max} \gets \texttt{ceil}(\texttt{max}(\texttt{norm}(P, \text{dim}=1)))$
\State $B_{\rho} \gets \texttt{arange}(-\rho_{\max}, \rho_{\max} + \Delta\rho, \Delta\rho)$
\State $\rho_{\text{idx}} \gets \texttt{bucketize}(\rho, B_{\rho})$

\Comment{Vote in the accumulator}
\State $A \gets \texttt{zeros}(\texttt{len}(B_{\rho}), \texttt{len}(\Theta))$
\State $\texttt{scatter\_votes}(A, \rho_{\text{idx}}, \Theta)$

\Comment{Smooth the accumulator}
\State $K \gets$ predefined kernel
\State $A \gets \texttt{conv2d}(A, K, \text{padding}=1)$

\Comment{Detect peaks in the accumulator}
\State $M \gets A > 0.3 \cdot \texttt{max}(A)$
\State $(i_r, i_\theta) \gets \texttt{nonzero}(M)$
\State $R, T \gets B_{\rho}[i_r], \Theta[i_\theta]$

\Comment{Assign points to each detected line}
\State $D \gets |R - (X \cos T + Y \sin T)|$
\State $M_{\text{assign}} \gets D < \Delta\rho$
\State $C \gets \texttt{sum}(M_{\text{assign}}, \text{dim}=1)$

\Comment{Return points from the line with most support}
\State $k^* \gets \texttt{argmax}(C)$
\State $P_{\text{max}} \gets P[M_{\text{assign}}[k^*]]$
\State \Return $P_{\text{max}}$
\end{algorithmic}
\end{algorithm}

\noindent\textit{Post-processing:}  
We extract ruler marks from the predicted heatmap using a custom PyTorch implementation of a local maxima detection approach, as detailed in Algorithm~\ref{alg:extract_points_onnx}. Hough line detection is also implemented in PyTorch and operates directly on the detected points, as shown in Algorithm~\ref{alg:hough_line_onnx}. All parameters for Algorithms~\ref{alg:extract_points_onnx} and~\ref{alg:hough_line_onnx} were selected heuristically based on manual inspection of the results of 10 sample images in the training dataset. Parameters were adjusted until the predictions were correct and the detected rulers appeared visually accurate for all 10 images. For clarity, these values appear as default parameters in the algorithm listings. For the computation of the Chamfer distance, we use a brute-force implementation with $\mathcal{O}(n^2)$ complexity. Differential evolution is implemented using scikit-learn's optimizer with the \texttt{rand1bin} strategy and Latin hypercube initialization.

The bounds for optimization are set as $r_{\max} = 1.5$, $r_{\min} = \frac{1}{1.5}$, $\mathcal{D}_{\max} = \max(\mathbf{D}_{\text{direct}})$, and $\mathcal{D}_{\min} = \min(\mathbf{D}_{\text{direct}})$, where $\mathbf{D}_{\text{direct}}$ is the set of distances between adjacent detected points. For DeepGP, no bounds are required, and all input mark locations are min-max normalized to the range \([-1, 1]\).

\subsection{Comparison with Existing Methods}
\revblue{We compared the final RulerNet-DeepGP system against representative digit-detection and mark-frequency baselines.}

\begin{table}[!t]
    \centering
    \footnotesize
        \caption{Performance comparison with existing methods (mAPE/cm@768). RulerNet was evaluated at a resolution of $768 \times 768$, while the Digits Detection~\citep{lukashchuk2022method} method was implemented by us and evaluated at the original resolution. The Mark Frequency~\citep{wen2024ruler} method was evaluated by the original authors at our request. Both the mean performance and standard deviation are reported (Mean$\pm$SD). For the inference time, milliseconds per \revblue{sample} (ms/sample) is computed as the time required for the method to extract the scale from the image on the AnyRuler dataset.  Results are reported on both AnyRuler and Rulers2023~\citep{matuzevivcius2023rulers2023} datasets.}
    \begin{tabular}{lccc}
    \toprule
         &  AnyRuler& Rulers2023 & ms/sample\\\\
         \midrule
        Digits Detection & 56.71\scalebox{0.8}{$\pm$92.23} & 76.05\scalebox{0.8}{$\pm$95.27} & 1530.62 \\
        Mark Frequency & 62.11\scalebox{0.8}{$\pm$134.41} & 14.79\scalebox{0.8}{$\pm$25.71} & - \\
         \midrule
        {\bf RulerNet-GP} & {1.32\scalebox{0.8}{$\pm$14.75}} & {1.46\scalebox{0.8}{$\pm$4.82}} & 2516.87\\
        {\bf RulerNet-DeepGP} & {\bf 1.19\scalebox{0.8}{$\pm$5.03}} & {\bf 1.31\scalebox{0.8}{$\pm$3.15}} & {\bf 42.18} \\
    \bottomrule
    \end{tabular}

    \label{tab:sotas}
\end{table}

Although previous methods have demonstrated strong performance in specific domains, Table~\ref{tab:sotas} shows that RulerNet is the only method capable of reliably performing ruler reading across a diverse range of images, achieving a pixel error of less than 1.5 pixels per cm. In contrast, existing approaches fail to generalize effectively in unconstrained settings. These results reinforce our motivation for the design of RulerNet and validate its robustness across real-world scenarios. Furthermore, RulerNet-DeepGP improves both accuracy and efficiency, enabling real-time ruler reading. Our final implementation is able to extract the ruler—i.e., the GP parameters—in an end-to-end fashion and is lightweight enough to be deployed on edge devices.

\begin{figure}
    \centering
    \includegraphics[width=\linewidth]{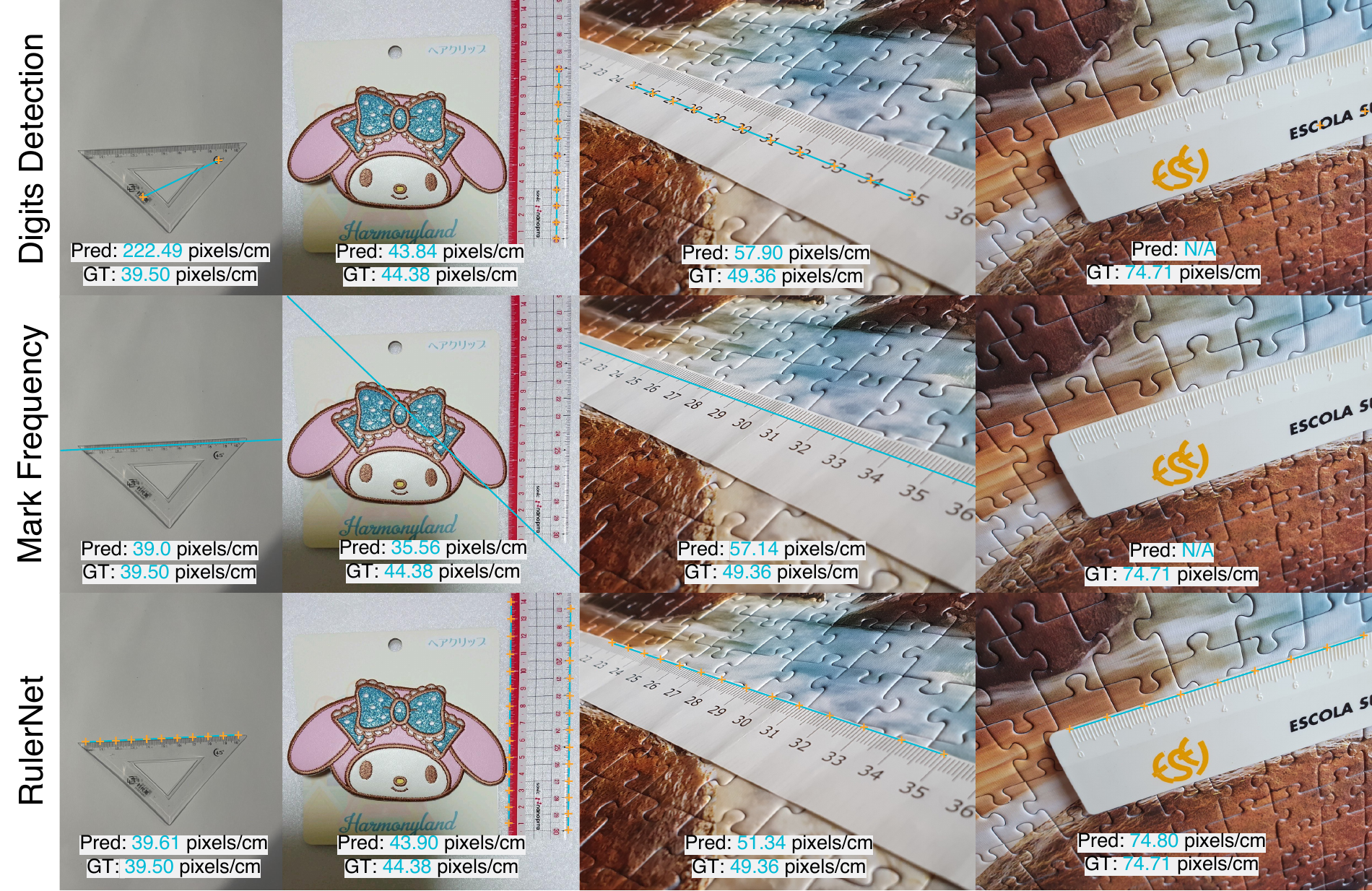}
    \caption{Qualitative examples from the AnyRuler and Rulers2023 datasets, showing ruler localization and the final mean pixels/cm estimation. Note that the Mark Frequency method does not detect individual marks but instead identifies a line along the marks and computes a single frequency value; therefore, the orange `+' markers are absent.}
    \label{fig:qual_sotas}
\end{figure}

From Fig.~\ref{fig:qual_sotas}, we observe that RulerNet handles challenging cases, such as when the mark color is similar to the ruler color or when there is significant perspective distortion. The larger error in the third column arises from averaging the pixels/cm across all cm segments in a distorted image, despite the detected marks themselves being highly accurate across all images. In contrast, the Digits Detection method fails primarily due to the OCR model's inability to reliably detect digits. When digits are detected, the subsequent GP module performs as expected. Because the GP module is not applied when no digits are detected, the Digits Detection method can be faster than RulerNet-GP despite using the same GP method. One might argue that tuning the OCR model on a ruler dataset with digit location annotations could yield performance comparable to that of RulerNet. However, such tuning is more resource-intensive, as it requires annotations for all digits rather than simple point marks, and it is inherently less robust---whereas not all rulers contain digits, all rulers have marks. For the Mark Frequency method, the blue lines represent the most confident line along the ruler marks, and errors occur both during ruler localization and frequency estimation. These observations further support the design of RulerNet, where direct mark localization produces more robust and generalizable results.

\subsection{Ablation Study}
In this subsection, we analyze the implications of our design choices and identify the components that contribute to a robust ruler reading model. All models were trained using the same architecture at an input resolution of $768\times768$, as larger training sizes would have incurred higher computational costs. \revblue{Specifically, Table~\ref{tab:abl_modules} quantifies the contributions of GP modeling and DeepGP relative to direct and median-based scale extraction, while Tables~\ref{tab:abl_graphic_number} and \ref{tab:abl_train_data} quantify the contributions of graphics-based and generative synthetic data.}

\noindent\textit{Why did we prompt ControlNet with ``rectangle'' instead of ``ruler''?}  

\begin{figure}
    \centering
    \includegraphics[width=0.95\linewidth]{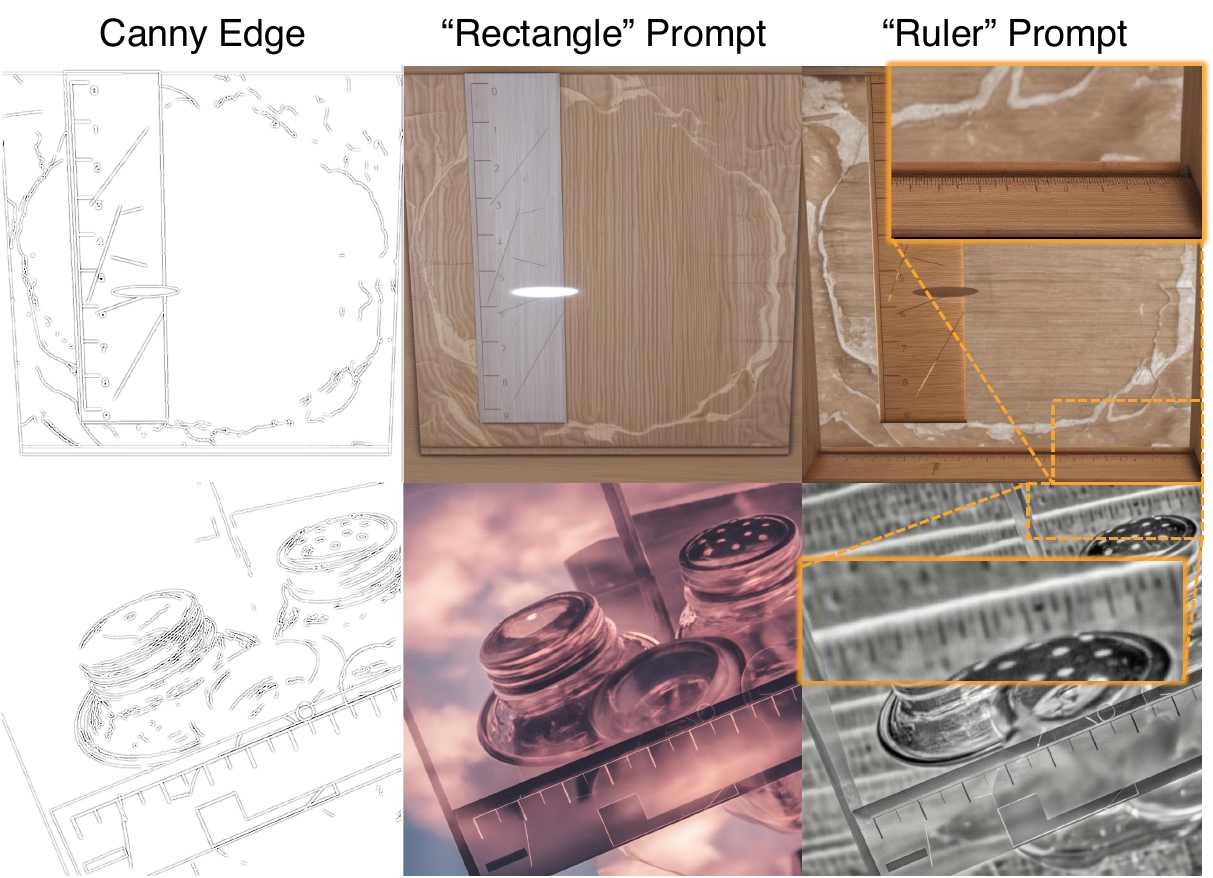}
    \caption{Comparison of the prompt words ``rectangle'' and ``ruler'' in generated ruler images. The Canny edge maps are inverted for improved visualization.}
    \label{fig:ruler_vs_rectangle}
\end{figure}

In our data generation pipeline, we deliberately used ``rectangle'' in the prompt instead of ``ruler''. Even with ControlNet, it is difficult to isolate the location of specific features. If we directly prompted for a ``ruler'', the model could have misinterpreted other parts of the image as rulers, leading to the generation of unintended marks. This phenomenon is akin to hallucinations observed in diffusion models~\citep{aithal2025understanding}. As shown in Fig.~\ref{fig:ruler_vs_rectangle}, using ``ruler'' as the prompt can result in erroneous ruler marks. By contrast, only the rulers generated from the graphics-based ruler generator included mark labels. Any unexpected marks generated in response to ``ruler'' would have been treated as background by the model, potentially degrading performance. Since our objective was to maintain full control over the ruler placement, we used the word ``rectangle'' instead of ``ruler''.

\noindent\textit{How did the ruler estimation method affect performance and speed?}

To evaluate the impact of the ruler/scale estimation method on both performance and inference speed, we compared four different approaches:
\begin{itemize}
    \item \textit{Direct adjacent points:} Compute the distances between all adjacent points and take the average.
    \item \textit{Median-based filtering:} Compute the ratios between adjacent distances, retain only those within a 20\% deviation from the median distance and a 10\% deviation from the median ratio, and then average over the remaining distances.
    \item \textit{GP optimization method:} Apply the proposed GP optimization technique to estimate the entire ruler and compute the scale.
    \item \textit{DeepGP:} Use the proposed DeepGP model to estimate the ruler's GP parameters and compute the scale.
\end{itemize}

\begin{figure}
    \centering
    \includegraphics[width=0.95\linewidth]{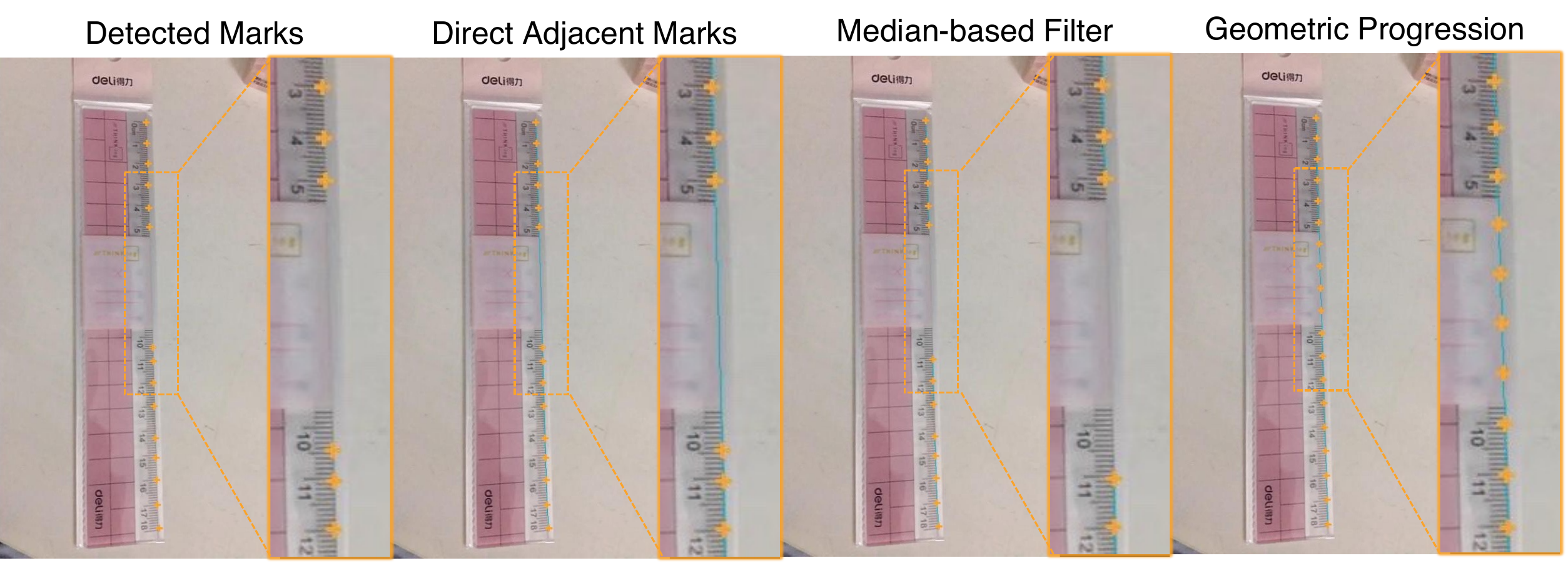}
    \caption{Qualitative comparison of different ruler scale estimation methods. Images are zoomed in for improved visualization.}
    \label{fig:ruler_mode}
\end{figure}

As shown in Table~\ref{tab:abl_modules}, directly averaging adjacent point distances yielded the worst performance, highlighting the effect of noise in the detected points and the need for further post-processing. The median-based filtering method significantly improved accuracy on both datasets by mitigating the influence of outliers. Applying the GP optimization method resulted in further improvements, particularly on the Rulers2023 dataset. As illustrated in Fig.~\ref{fig:ruler_mode}, this method not only recovered the full ruler but also corrected inaccurate detections, offering finer control over scale estimation.

More importantly, the proposed DeepGP model achieved additional performance gains across both datasets, suggesting that the training data generation strategy described in Algorithm~\ref{alg:noisy_gp} produced representative and diverse training samples. \revblue{This performance gap is consistent with the motivation for DeepGP described in the Methods section. An optimization-based GP fit might appear to provide an upper bound under ideal noise-free constraints; however, Table~\ref{tab:abl_modules} evaluates the methods on RulerNet detector outputs, where the input points can be noisy, missing, spurious, or slightly shifted. Under these realistic detector errors, the bounded optimizer is not a true oracle because it directly fits the noisy point set and can be pulled toward incorrect adjacent intervals. DeepGP was designed specifically for this setting: it learns a data-driven prior over plausible ruler sequences and jointly predicts validity, adjacency, and point offsets, allowing it to denoise the RulerNet detections before scale estimation. Thus, the improvement from 2.20 to 1.81 mAPE/cm@768 is consistent with our design of enabling DeepGP to better handle noisy detected marks than direct optimization-based GP fitting.} In addition to improved accuracy, DeepGP delivered over a 100$\times$ speedup in inference time compared to the optimization-based GP method, while maintaining comparable speed to other less generalizable approaches. This efficiency is achieved by leveraging GPU acceleration through a lightweight, feed-forward design.

\begin{table}[!t]
    \centering
    \footnotesize
   \caption{Evaluation of RulerNet with different scale estimation methods (mAPE/cm@768), trained solely on the AnyRuler dataset. Milliseconds per \revblue{sample} (ms/sample) is computed as the average time it takes to extract the ruler from the predicted marks on the AnyRuler dataset. Results are reported on both AnyRuler and Rulers2023~\citep{matuzevivcius2023rulers2023} datasets.}
    \begin{tabular}{lcccc}
    \toprule
         &  AnyRuler & Rulers2023 & Avg. & ms/sample\\
         \midrule
        RulerNet-Direct & 1.53 & 4.43 & 2.98 & 16.99\\
        RulerNet-Median & 1.33 & 3.65 & 2.49 & 16.08\\
        RulerNet-GP & 1.38 & 3.02 & 2.20 & 2497.33\\
        RulerNet-DeepGP & 0.84 & 2.78 & 1.81 & 22.61\\
    \bottomrule
    \end{tabular}
    \label{tab:abl_modules}
\end{table}

\noindent\textit{How did image resolution affect performance?} 

To study the effect of input image resolution on the model, we tested four different inference input resolutions using the baseline RulerNet trained on AnyRuler, with direct adjacent-point distances used for scale estimation. Results were reported on both the AnyRuler and Rulers2023 testing sets.

As shown in Table~\ref{tab:abl_resolution}, increasing resolution generally improved performance; however, beyond a certain point, further increases yielded diminishing returns. Notably, we observed performance degradation beginning at 1,024 on AnyRuler, while performance on Rulers2023 plateaued after 1,024. Further investigation revealed that AnyRuler images had a lower original resolution (an average of 1,036, with only 47\% of images exceeding 768) compared to Rulers2023 (an average of 3,528, with all images exceeding 768). This suggests that AnyRuler may not benefit as much from higher input resolutions.

\begin{figure}
    \centering
    \includegraphics[width=0.95\linewidth]{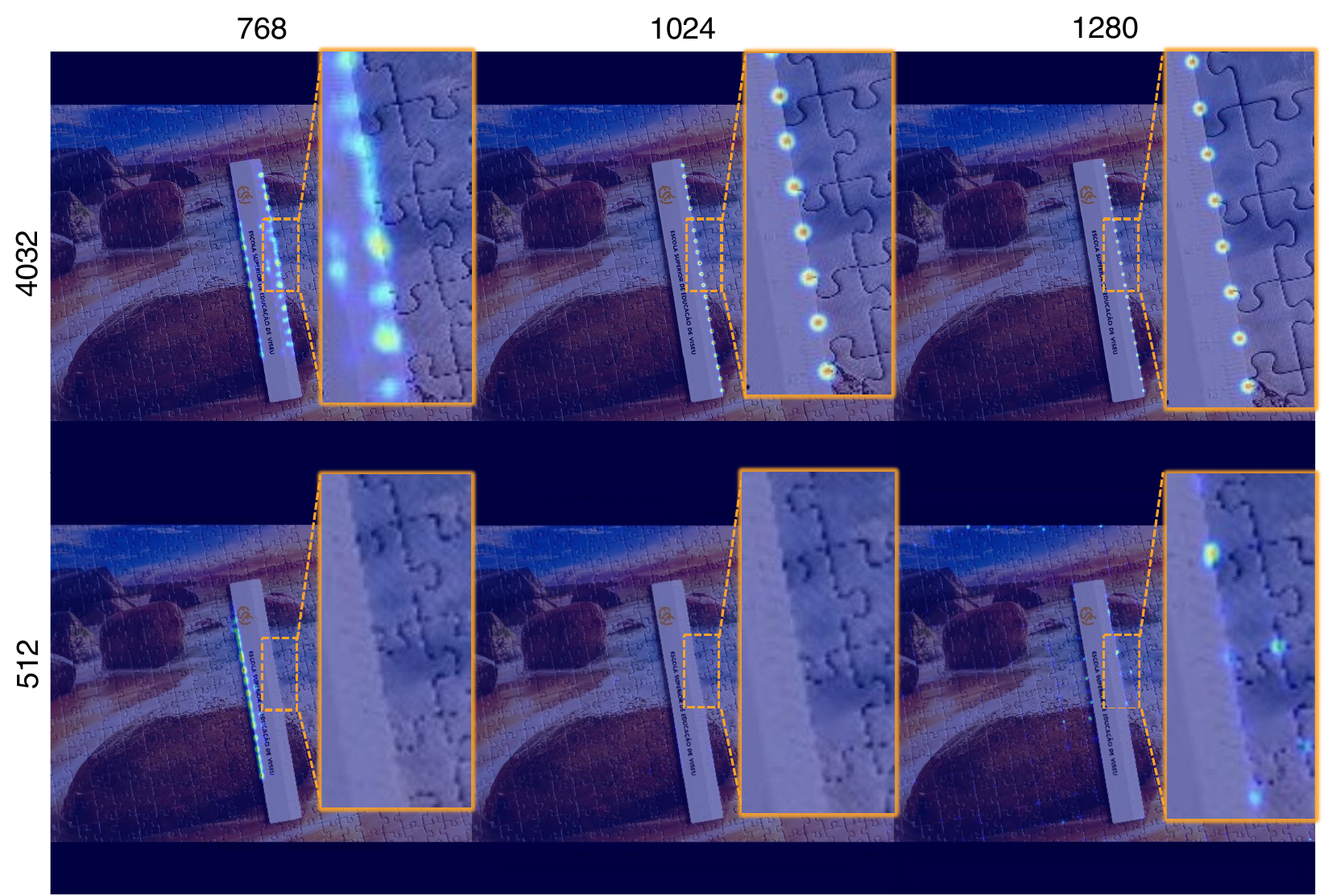}
    \caption{Qualitative example illustrating the effect of inference input resolution. Each column represents a different input resolution to the model. Each row represents a different original image resolution. Images are zoomed in for improved visualization.}
    \label{fig:abl_resolutions}
\end{figure}

Additionally, we visualized the output heatmap at high and low resolutions, and for low-resolution images upsampled to high resolution (see Fig.~\ref{fig:abl_resolutions}). The observations indicate that when the original resolution is insufficient for the marks to be clearly visible (e.g., some images in the AnyRuler dataset), upsampling does not improve performance. This further supports our numerical findings. 

\begin{table}[!t]
\centering
\footnotesize
\caption{Evaluation of RulerNet-Direct on different inference resolutions (mAPE/cm@768). The model is trained solely on the AnyRuler dataset. Results are reported on both AnyRuler and Rulers2023~\citep{matuzevivcius2023rulers2023} datasets.}
\begin{tabular}{lccc}
\toprule
     &  AnyRuler & Rulers2023 & Avg.\\
     \midrule
    512 & 2.88 & 9.77 & 6.33\\
    768 & 1.53 & 4.43 & 2.98\\
    1,024 & 1.76 & 2.82 & 2.29\\
    1,280 & 1.82 & 2.85 & 2.34\\
\bottomrule
\end{tabular}
\label{tab:abl_resolution}
\end{table}

\noindent\textit{How did the number of synthetic rulers per training epoch affect performance?} 

To study the effect of the number of synthetic rulers in each training epoch, in addition to using the AnyRuler dataset, we gradually increased the number of synthetic rulers up to 2,000 images until performance plateaued or began to decline. Since we have two types of synthetic rulers, we analyzed them in isolation; the combined effect is addressed later.

As shown in Table~\ref{tab:abl_graphic_number}, incorporating more synthetic rulers improved generalizability on the Rulers2023 dataset. Specifically, using generative images tended to enhance generalizability more than using graphics-based rulers. Although minor fluctuations in performance on the AnyRuler dataset were observed with generative rulers, the overall gains in generalization justified their increased use. Therefore, employing more generative synthetic rulers per training epoch improved model performance.

\begin{table}[!t]
\centering
\footnotesize
\caption{Evaluation with different numbers of graphics-based or generative rulers during training (mAPE/cm@768) using the direct adjacent points distance.}
\begin{tabular}{lcccc}
\toprule
    &\multicolumn{2}{c}{Graphics-based} & \multicolumn{2}{c}{Generative} \\
     \cmidrule(lr){2-3}\cmidrule(lr){4-5}
     &  AnyRuler & Rulers2023 &  AnyRuler & Rulers2023\\
     \midrule
    100 & 1.70 & 4.43 & 1.67 & 4.36\\
    1,000 & 1.55 & 4.06 & 1.58 & 3.56 \\
    2,000 & 1.55 & 3.48 & 1.64 & 2.94 \\
\bottomrule
\end{tabular}
\label{tab:abl_graphic_number}
\end{table}

\noindent\textit{How did the use of different synthetic ruler types affect performance?} 

We investigated how the combination of two different types of synthetic rulers affected model performance. First, we evaluated each type individually by training with 1,000 synthetic rulers from each method as a reference. Then, we evaluated the effect of using both types together during training.

As shown in Table~\ref{tab:abl_train_data}, incorporating additional synthetic rulers improved performance on the Rulers2023 dataset. When combining both graphics-based and generative rulers, the improvement in generalizability on Rulers2023 was much greater than the slight performance degradation observed on AnyRuler, suggesting a potential trade-off between in-domain performance and generalizability. Based on results reported earlier, where adding more graphics-based rulers improved performance on both datasets, we argue that this perceived trade-off may result from insufficient training on a more diverse dataset. \revblue{To test this hypothesis, we pre-trained the model using the same CE+DICE heatmap objective on the AnyRuler training set together with both synthetic sources: graphics-based rulers generated on the fly by the programmatic generator and 102,000 fixed, pre-generated ControlNet-refined generative rulers. We then trained the model with the AnyRuler training set and both synthetic sources. In both pre-training and training, graphics-based rulers were generated on the fly, whereas generative rulers were pre-generated. The only difference among the rows in Table~\ref{tab:abl_train_data} is the number and composition of synthetic data sampled from these two sources.} The number of synthetic rulers was chosen to balance performance and computational cost. This approach outperformed all previous configurations on both datasets. These results indicate that the additional use of synthetic rulers can greatly improve model performance, provided that the model is well-trained.

\begin{table}[!t]
    \centering
    \footnotesize
    \caption{Evaluation of the effect of combining graphics-based (Gra.) and generative (Gen.) rulers during training (mAPE/cm@768) using the direct adjacent points distance. ``Pre.'' indicates pre-training the model with a large number of graphics-based and generative rulers. Results are reported on both AnyRuler and Rulers2023~\citep{matuzevivcius2023rulers2023} datasets.}
    \begin{tabular}{lcc}
    \toprule
         &  AnyRuler & Rulers2023\\
         \midrule
        RulerNet & 1.53 & 4.43 \\
         + Gra. & 1.55 & 4.06 \\
         + Gen. & 1.58 & 3.56 \\
         + Gen. + Gra. & 1.61 & 2.87 \\
         + Gen. + Gra. + Pre. & 1.50 & 1.82 \\
    \bottomrule
    \end{tabular}
    \label{tab:abl_train_data}
\end{table}

These ablation studies further validate the effectiveness of the proposed data generation pipeline, model training strategy, and post-processing method. With proper settings, RulerNet achieves excellent performance.

\noindent\textit{\revblue{How robust are DeepGP and the detector to noisy detections and mark confusion?}}

\begin{figure}
    \centering
    \includegraphics[width=0.98\linewidth]{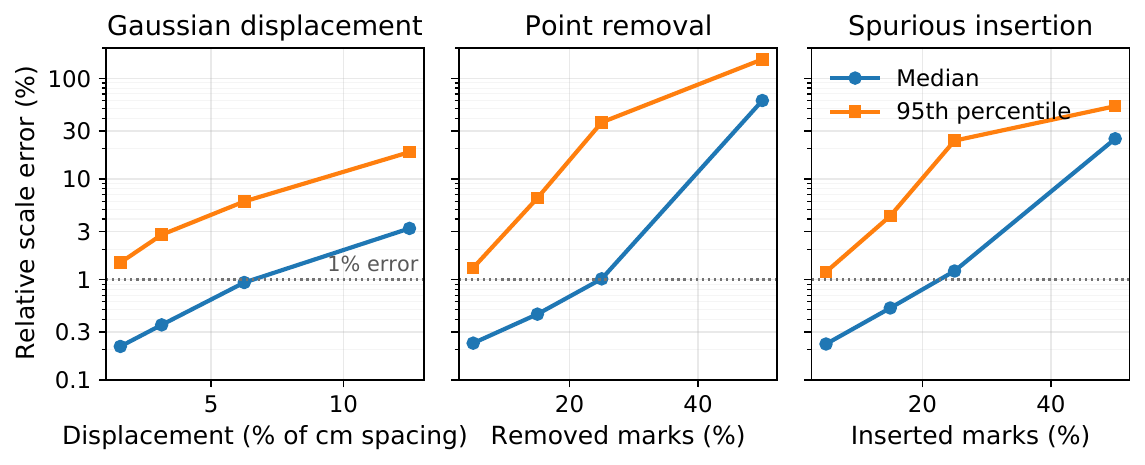}
    \caption{\revblue{DeepGP robustness to controlled mark-detection noise. Perturbations are applied to detected ruler marks before DeepGP inference, isolating the stability of the learned GP recovery module. Curves show median and 95th-percentile relative scale error for each corruption type on a log-scaled y-axis; the dotted horizontal line marks 1\% relative scale error. The combined moderate-noise condition produced 0.49\% median and 3.73\% 95th-percentile relative scale error.}}
    \label{fig:deepgp_noise_curves}
\end{figure}

\revblue{We first evaluated DeepGP by perturbing detected ruler marks before DeepGP inference. Fig.~\ref{fig:deepgp_noise_curves} shows that DeepGP is robust under moderate detector-output corruption: Gaussian displacement up to 6.25\% of local spacing, 15\% point removal, and 15\% spurious insertion each produced sub-1\% median scale change. Under a combined moderate-noise condition with 0--15\% point removal and insertion together with small positional jitter, the estimated scale changed by a median of 0.49\%, with a 95th-percentile change of 3.73\%. The 50\% point-removal and 50\% spurious-insertion settings represent intentionally extreme out-of-distribution stress cases rather than realistic imaging conditions; they are included to illustrate the point at which the learned GP recovery begins to break down.}

\revblue{We then tested whether RulerNet confuses cm and mm marks. On all 367 annotated AnyRuler rulers, we matched each predicted point to an annotated cm mark using a 1 mm tolerance, i.e., 10\% of the local 1 cm spacing. A matched prediction was counted as a true positive, an unmatched annotated cm mark as a false negative, and an unmatched prediction as a false positive. Under this criterion, RulerNet achieved a 90.28\% one-to-one precision and an 85.19\% one-to-one recall (Table~\ref{tab:cm_mm_consistency}). The average predicted mark count difference was $-5.46\%$, indicating that RulerNet usually predicts slightly fewer marks than were annotated, rather than adding extra marks. Together with the high one-to-one precision, this suggests that RulerNet does not usually confuse mm marks with cm marks, although it may miss some cm marks. Because the DeepGP analysis above demonstrates robustness to 15\% point removal, 15\% spurious insertion, and small localization noise, these minor detector errors are unlikely to materially affect the final scale estimate after DeepGP correction.}

\begin{table}[!t]
    \renewcommand{\tabcolsep}{3pt}
    \footnotesize
    \begin{center}
    \begin{tabular}{lc}
        \toprule
        \revblue{Metric} & \revblue{Value} \\
        \midrule
        \revblue{Average count difference} & \revblue{$-5.46\%$} \\
        \revblue{One-to-one precision} & \revblue{90.28\%} \\
        \revblue{One-to-one recall} & \revblue{85.19\%} \\
        \bottomrule
    \end{tabular}
    \end{center}
    \caption{\revblue{Cm-mark count and localization consistency on the AnyRuler test set. Because mm marks are not separately annotated, this table reports average cm-mark count difference and one-to-one localization agreement rather than direct mm false-positive labels.}}
    \label{tab:cm_mm_consistency}
\end{table}

\noindent\textit{\revblue{How does RulerNet behave under controlled perspective and image-quality stressors?}}

\begin{figure}
    \centering
    \includegraphics[width=0.98\linewidth]{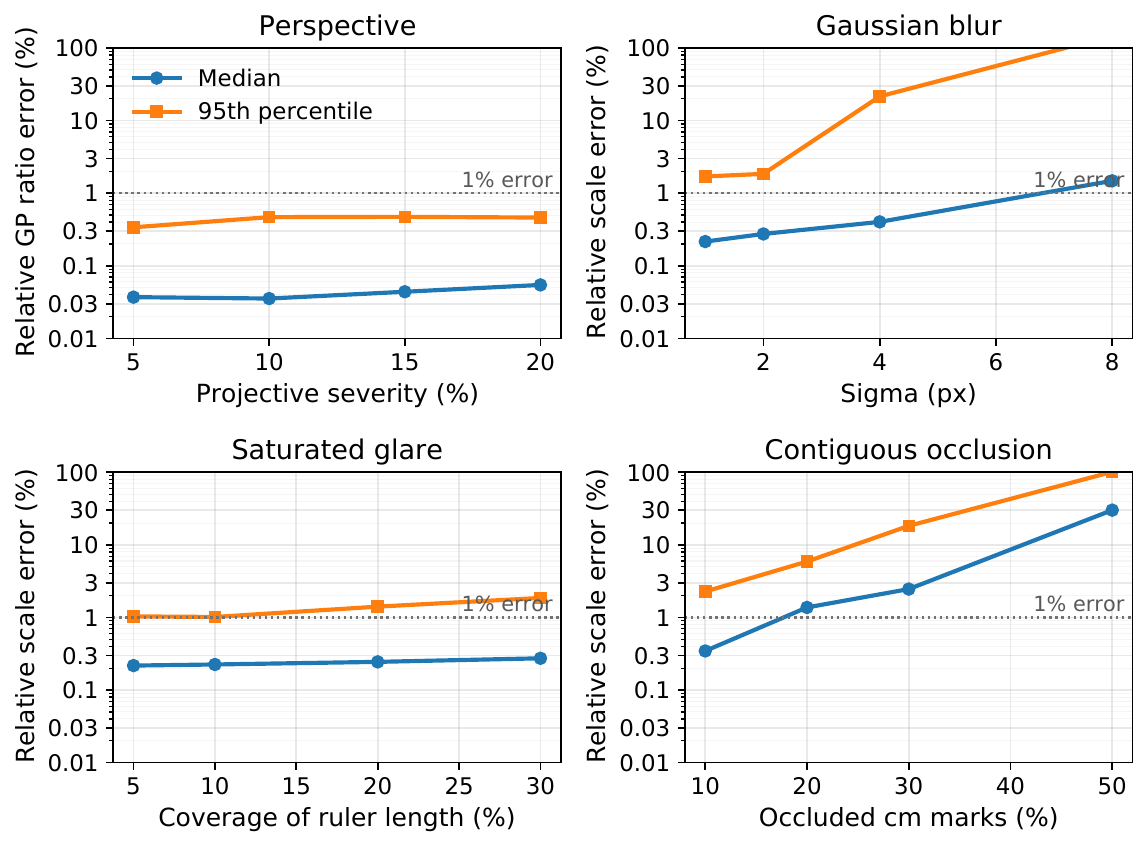}
    \caption{\revblue{Controlled perspective and image-quality stress tests on 340 clean-accurate AnyRuler ruler instances. The perspective panel reports median and 95th-percentile relative GP-ratio error between the predicted ratio and the ratio induced by the transformed ground-truth marks. The image-quality panels report median and 95th-percentile relative scale error for successful predictions. All panels use log-scaled y-axes.}}
    \label{fig:perspective_quality_stress_curves}
\end{figure}

\revblue{To test whether the full RulerNet+DeepGP pipeline is robust to perspective-transformed rulers, we selected 340 AnyRuler instances with accurate baseline predictions, defined as having a DeepGP scale error of at most 10\% and at least three detected marks. We then applied controlled homographies jointly to each image and its cm-mark annotations, so that the ground-truth scale and GP ratio were transformed consistently with the image. At the strongest projective-transformation severity of 20\%, the transformed cm intervals had a median coefficient of variation of 17.07\% and a median longest-to-shortest interval ratio of 1.75. We directly compared the predicted GP ratio with the ratio induced by the transformed ground-truth marks. At this severity, the median relative GP-ratio error was 0.055\%, and the 95th-percentile error was 0.463\% (Fig.~\ref{fig:perspective_quality_stress_curves}). These results suggest that RulerNet+DeepGP remains robust under a wide range of planar perspective transformations. This test demonstrates the consistency of the proposed GP representation under known planar perspective distortion; however, it does not model curved rulers or depth differences between the ruler and the object.}

\begin{figure}
    \centering
    \includegraphics[width=0.98\linewidth]{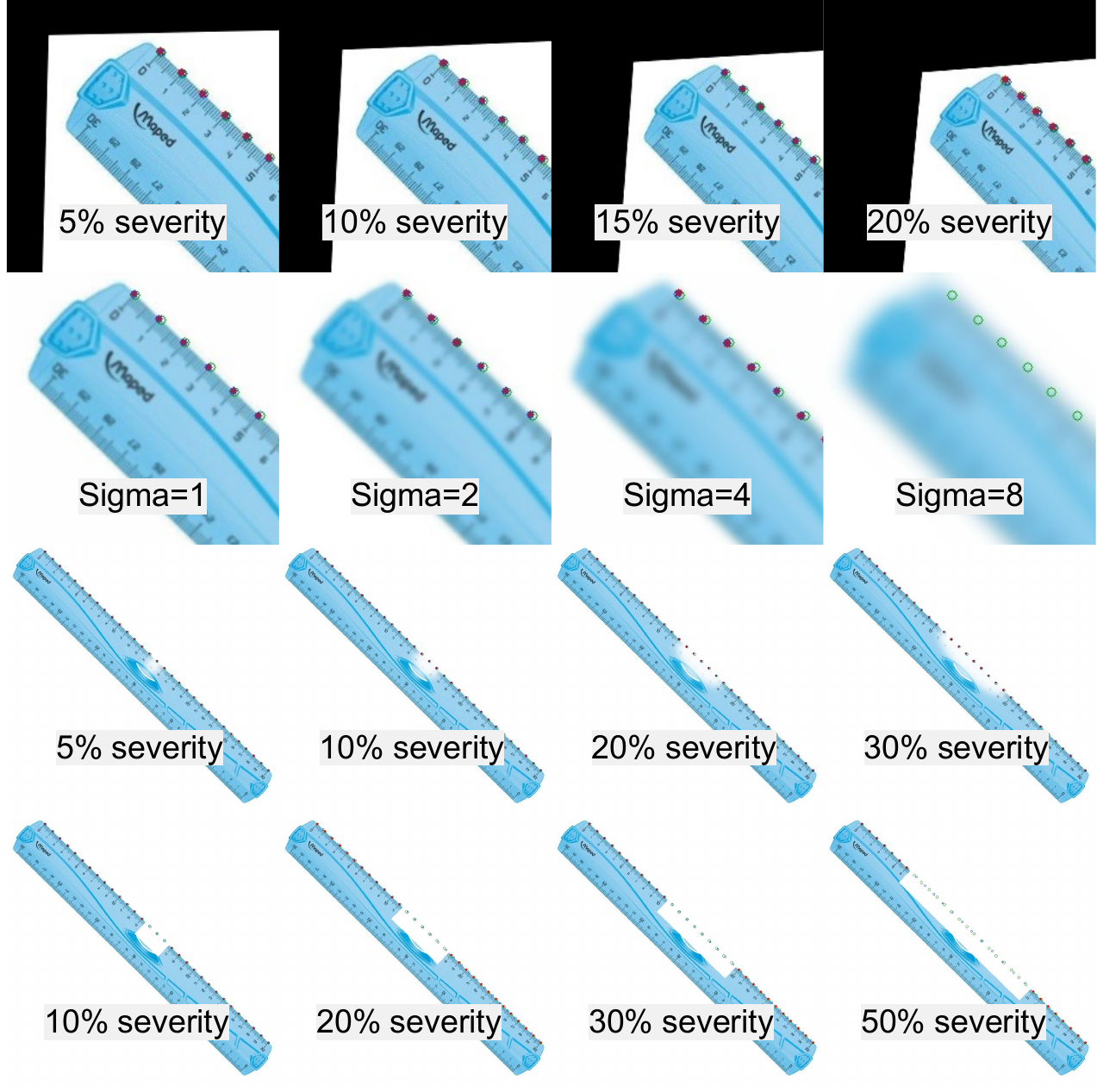}
    \caption{\revblue{Example visualizations of the controlled stressors at increasing severity, rendered from the original-resolution source image for visual clarity. Rows show perspective transformation, Gaussian blur, saturated glare, and contiguous occlusion; columns show increasing severity for each stressor. The quantitative stress-test results are computed after the standard $768\times768$ validation preprocessing described in the text.}}
    \label{fig:controlled_stress_strength_examples}
\end{figure}

\revblue{Using the same 340 AnyRuler instances with accurate baseline predictions, we also evaluated controlled blur, glare, and contiguous occlusion. Each image was first padded and resized using the standard validation preprocessing to the $768\times768$ model input coordinate system; the quality corruptions were then applied in this fixed coordinate system. This ensures that the blur parameter, glare coverage, and occlusion coverage are comparable across images and remain aligned with the transformed cm-mark annotations. This stress test complements broader image-quality assessment work: instead of predicting perceptual image quality, it directly measures how specific degradations affect scale estimation. Fig.~\ref{fig:controlled_stress_strength_examples} visualizes the different degradation levels, and Fig.~\ref{fig:perspective_quality_stress_curves} reports the corresponding scale-estimation errors. The model tolerated targeted saturated glare spanning up to 30\% of the ruler length, with a 0.27\% median relative scale error. Mild Gaussian blur was similarly robust at $\sigma=2$ pixels, with a 0.27\% median relative scale error. The degraded cases with $\sigma=4$ blur and 50\% contiguous mark occlusion represent intentionally extreme stress tests rather than realistic operating conditions; they are included to characterize the operating limits of the proposed method. Under these extreme settings, $\sigma=4$ blur increased the 95th-percentile relative scale error to 21.59\%, and 50\% mark occlusion produced 30.14\% median relative scale error.}

\subsection{Medical Pipeline Integration Evaluation}
As a scale estimation module, RulerNet is designed to operate in conjunction with downstream analysis methods to estimate the physical size of objects of interest. To evaluate RulerNet as a drop-in replacement for existing medical scale estimation pipelines, we replaced the scale estimation component in the AI-PLAX~\citep{chen2020ai} placenta analysis framework (part of the PlacentaVision project) with RulerNet and evaluated it on the same dataset used to tune AI-PLAX~\citep{chen2020ai}, collected at Northwestern Memorial Hospital. All rulers in the dataset are non-conventional printed rulers with only cm marks in the middle of the ruler. We adopt the long-axis length measured during gross examination as the ground truth for assessing measurement accuracy. Placenta images exhibiting accessory lobes, bilobed morphology, folding, an incomplete disc, or missing rulers were manually excluded, as gross examination staff typically reach consensus only for well-formed placentas, resulting in a total of 711 images. We then assessed final long-axis length prediction accuracy using the mean absolute error (MAE), directly comparing our method against the original AI-PLAX~\citep{chen2020ai} pipeline. To isolate scale estimation errors, we used the long axis detected by AI-PLAX~\citep{chen2020ai} for both methods. As shown in Table~\ref{tab:plax_rulernet}, RulerNet achieves a lower MAE than the baseline method despite not being tuned on this dataset, suggesting that RulerNet provides a more robust scale estimation component.

\begin{table}[!t]
    \centering
    \footnotesize
    \caption{Comparison of long-axis length estimation error (cm). Mean absolute error (MAE) and standard deviation (SD) are reported as Mean$\pm$SD.}
    \begin{tabular}{lc}
        \toprule
        Method & MAE  \\
        \midrule
        AI-PLAX & 1.68\scalebox{0.8}{$\pm$}4.27 \\
        \midrule
        {\bf RulerNet} & {\bf 1.16\scalebox{0.8}{$\pm$}1.01} \\
        \bottomrule
    \end{tabular}
    \label{tab:plax_rulernet}
\end{table}

\begin{figure}
    \centering
    \includegraphics[width=\linewidth]{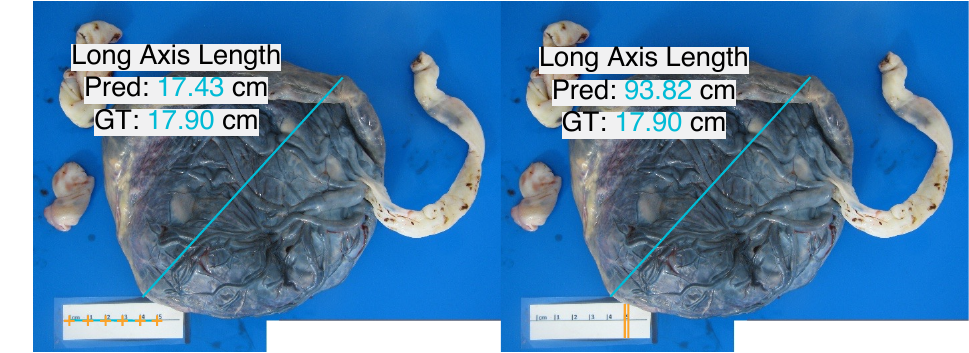}
    \caption{Qualitative comparison between RulerNet and AI-PLAX on placenta images. 
    Orange `+' markers denote RulerNet predictions; orange `|' markers denote AI-PLAX predictions; the blue line indicates the detected long axis of the placenta. }
    \label{fig:intergration}
\end{figure}

We performed additional visual analysis to identify where AI-PLAX~\citep{chen2020ai} introduces errors. As shown in Fig.~\ref{fig:intergration}, most AI-PLAX~\citep{chen2020ai} failures occur when numbers printed next to ruler marks are incorrectly identified as marks themselves. This behavior is expected, as AI-PLAX~\citep{chen2020ai} relies on kernel density estimation to detect ruler marks and therefore cannot reliably distinguish between true tick marks and adjacent numeric labels. These results further emphasize the importance of avoiding fragile handcrafted features; although a method can be tuned to a specific application, such solutions often fail to generalize.

\section{Limitations and Future Directions}

Although RulerNet achieves state-of-the-art performance and robustness for ruler reading in the wild, several limitations constrain its current scope and point to avenues for future research. 

\begin{itemize}
\item \noindent\textit{\revblue{Metric-ruler scope and unit adaptation:}} \revblue{The current RulerNet model is optimized for metric rulers with cm marks; inch-only rulers were excluded from the primary AnyRuler benchmark, and inch marks were treated as non-target marks during training. Thus, the released model should not be interpreted as a zero-shot inch-ruler reader. To test adaptation to inch-based rulers, we generated ControlNet-based synthetic inch-only rulers and trained the model on 800 images with 200 held out for evaluation. This adaptation produced no scale-estimation failures and achieved a mean error of 0.242 pixels/cm and a median per-image error of 0.065 pixels/cm, suggesting that inch support requires additional data and training rather than changes to the underlying framework.}

\item \noindent\textit{Planarity and flexibility:} Our approach assumes that the measuring tool is at least locally straight, with perspective distortion modeled via the GP formulation. Flexible tools such as tape measures are supported only when they are laid out in a straight configuration; measurements taken along curved, bent, or deformed shapes fall outside the scope of this work. While precalibration can correct for global effects (e.g., radial lens distortion~\citep{zhang2011camera}), extending the method to handle non-straight or curved measuring tools would require explicit modeling of non-planar geometry or adaptive distortion compensation.

\item \noindent\textit{2D-to-3D generalization:} RulerNet estimates scale in the 2D image plane, assuming that the ruler and the object lie on the same plane. Extending this approach to 3D would require estimating relative depth (e.g., via monocular depth prediction), but this is challenging due to the lack of RGB-D datasets annotated with ruler information, which complicates both method design and evaluation. Future work could address this limitation by collecting such datasets for systematic evaluation.

\item \noindent\textit{Out-of-plane distortion:} The GP module models perspective along the ruler’s length but not orthogonal distortion. Scale errors arise for objects offset from the ruler plane. Potential solutions include multi-ruler triangulation, inertial data fusion (e.g., gyroscope-based pose compensation~\citep{li2024gyroflow+}), or geometric rectification.

\item \noindent\textit{Ruler-mark detection robustness:} Both the Hough and GP modules depend on accurate ruler-mark detection, which can fail under occlusion, blur, or unusual markings. Future work could add learned uncertainty estimation or \revblue{an image-quality assessment module~\citep{wang2026ksiqa,luo2025macbiqa,wang2024aga,liu2022liqa,sun2022graphiqa}} to flag low-confidence outputs and prompt user guidance in difficult cases. 
\end{itemize}

\section{Conclusion}

In this work, we introduced RulerNet, a robust ruler-reading framework that integrates deep learning with geometric reasoning for scale estimation in the wild. \revblue{The method centers on cm-mark keypoint detection and GP-based scale recovery: the GP model compactly approximates the non-uniform spacing induced by perspective projection, while DeepGP efficiently recovers the underlying ruler structure from noisy, missing, or spurious detections. Together with the hybrid synthetic-data pipeline, which combines on-the-fly graphics-based generation with pre-generated ControlNet-refined rulers, this design improves generalization across diverse ruler appearances and imaging conditions without requiring full camera calibration.}

More broadly, this work demonstrates how AI-driven measurement systems can bridge the gap between digital perception and physical dimensions. In addition to meeting our internal needs as part of the PlacentaVision project—where accurate measurements support morphological characterization of the placenta~\citep{chen2020ai} and clinical assessment of maternal and neonatal health~\citep{pan2024cross}—our method addresses the growing demand for scalable and adaptable measurement tools in fields such as augmented reality, robotics, cataloging, and medical diagnostics. RulerNet offers a practical and extensible blueprint for integrating scale estimation with segmentation and scene understanding methods. By minimizing the need for manual annotations and leveraging both graphics- and diffusion-based synthetic data, RulerNet paves the way for the next generation of AI-assisted measurement tools—systems that are not only accurate but also generalizable, efficient, and ready for deployment in real-world scenarios.

```latex
\section*{CRediT authorship contribution statement}

\textbf{Yimu Pan:} Writing -- review \& editing, Writing -- original draft, Visualization, Validation, Software, Resources, Methodology, Investigation, Formal analysis, Data curation, Conceptualization.
\textbf{Manas Mehta:} Validation, Software, Data curation, Conceptualization.
\textbf{Gwen Sincerbeaux:} Writing -- review \& editing, Data curation.
\textbf{Jeffery A. Goldstein:} Writing -- review \& editing, Funding acquisition, Conceptualization.
\textbf{Alison D. Gernand:} Writing -- review \& editing, Funding acquisition, Supervision, Project administration, Conceptualization.
\textbf{James Z. Wang:} Writing -- review \& editing, Supervision, Resources, Funding acquisition, Conceptualization.

\section*{Declaration of competing interest}

The authors declare the following financial interests or personal relationships which may be considered as potential competing interests: Alison Gernand reports financial support was provided by National Institutes of Health. Alison Gernand reports financial support was provided by Burroughs Wellcome Fund. James Wang reports equipment, drugs, or supplies was provided by ACCESS-CI, National Science Foundation. James Wang reports financial support was provided by National Institutes of Health. Jeffery Goldstein reports financial support was provided by National Institutes of Health. James Wang reports a relationship with The Pennsylvania State University that includes employment. James Wang has patent \textit{SYSTEM AND METHOD FOR ROBUST RULER READING IN THE WILD} pending and assigned to The Pennsylvania State University. James Wang is an Associate Editor for the journal. He was not involved in the peer review process of this article. The other authors declare that they have no known competing financial interests or personal relationships that could have appeared to influence the work reported in this paper.

\section*{Acknowledgments}

Research reported in this publication was supported by the National Institute of Biomedical Imaging and Bioengineering of the National Institutes of Health (NIH) under award number R01EB030130 and by the Burroughs Wellcome Fund under award number 1271124. This project was 80\% supported by federal funds, with the federal award totaling \$2,008,840, and 20\% supported by non-federal funds, with the non-federal award totaling \$500,000.

The NIH award supported the development and validation of AI methods for analyzing digital placental photographs, including the RulerNet method reported here. The BWF award supported the development of PlacentaVision, an AI-driven mobile tool for placental imaging and the prediction of neonatal sepsis, including the image scale-estimation component described in this work.

This work used cluster computers at the National Center for Supercomputing Applications and the Pittsburgh Supercomputing Center through an allocation from the Advanced Cyberinfrastructure Coordination Ecosystem: Services \& Support (ACCESS) program, which is supported by National Science Foundation (NSF) Grants Nos. 2138259, 2138286, 2138307, 2137603, and 2138296. These resources supported the training and evaluation of the models reported in this study.

The content is solely the responsibility of the authors and does not necessarily represent the official views of the funding agencies. The authors thank the anonymous reviewers for their constructive comments, which helped improve the quality of this manuscript. The authors also acknowledge the encouragement and support of the other members of the PlacentaVision project team.



\bibliographystyle{cas-model2-names}

\bibliography{cas-refs}


\newpage
\appendix

\section{\revblue{GP Approximation to Projective Ruler Spacing}}
\label{app:gp_projective_approx}

{
This section provides a more detailed justification for using a geometric progression (GP) to model image-space ruler spacing under perspective. The key point is that GP is not an exact invariant of perspective projection; instead, it is a compact low-dimensional approximation to the smoothly varying spacing produced by projective geometry. This approximation is accurate over the finite ruler lengths and camera distances encountered in close-range imaging, while being much easier to fit robustly from noisy detected marks than a full projective camera model.

\noindent\textit{Physical ruler model.}
A physical ruler is assumed to be straight, and its centimeter marks are equally spaced in physical space. Let $x_i$ denote the one-dimensional physical coordinate of the $i$th centimeter mark along the ruler. Then
\begin{equation}
    x_i = x_0 + i\ell\;,
    \label{eqn:app_physical_spacing}
\end{equation}
where $\ell$ is the true physical spacing between adjacent centimeter marks. This is the only equal-spacing assumption: it applies before imaging, not after projection into the image.

\noindent\textit{Perspective projection along a straight line.}
A perspective projection maps a 3D line to an image line, provided the line does not pass through the camera center. Therefore, once the ruler is assumed to be straight, the projected mark sequence can be described by a one-dimensional projective transformation along the projected image line:
\begin{equation}
    u_i = \frac{a x_i + b}{c x_i + d}\;,
    \label{eqn:app_1d_homography}
\end{equation}
where $u_i$ is the scalar coordinate along the projected ruler line, $ad-bc\neq 0$, and $c x_i+d\neq 0$ for all visible marks. Choosing the orientation of $u_i$ so that the visible marks are ordered increasingly, the adjacent image-space interval is
\begin{align}
    \Delta_i^{\mathrm{true}}
    &= u_{i+1} - u_i \nonumber\\
    &=
    \frac{a x_{i+1}+b}{c x_{i+1}+d}
    -
    \frac{a x_i+b}{c x_i+d} \nonumber\\
    &=
    \frac{(ad-bc)(x_{i+1}-x_i)}{(c x_i+d)(c x_{i+1}+d)} \nonumber\\
    &=
    \frac{(ad-bc)\ell}{(c x_i+d)(c x_{i+1}+d)}\;.
    \label{eqn:app_true_delta}
\end{align}
Equivalently, if the opposite image-line orientation is used, $\Delta_i^{\mathrm{true}}$ can be interpreted as the absolute Euclidean interval $|u_{i+1}-u_i|$. Equation~\ref{eqn:app_true_delta} shows why equal physical spacing does not imply equal image spacing: the denominator changes with $i$ whenever $c\neq 0$. The affine case is recovered when $c=0$, in which case all intervals are multiplied by the same scale factor.

\noindent\textit{Why exact projected spacings are not a GP.}
Define the normalized perspective strength
\begin{equation}
    \eta = \frac{c\ell}{c x_0+d}\;.
    \label{eqn:app_eta}
\end{equation}
Because $x_i=x_0+i\ell$, we have
\begin{equation}
    c x_i+d=(c x_0+d)(1+i\eta)\;.
\end{equation}
Normalizing the projected spacings by the first interval gives
\begin{equation}
    D_i^{\mathrm{true}}
    =
    \frac{\Delta_i^{\mathrm{true}}}{\Delta_0^{\mathrm{true}}}
    =
    \frac{1+\eta}{(1+i\eta)(1+(i+1)\eta)}\;.
    \label{eqn:app_true_projective_spacing}
\end{equation}
The ratio of two consecutive normalized intervals is then
\begin{equation}
    \frac{D_{i+1}^{\mathrm{true}}}{D_i^{\mathrm{true}}}
    =
    \frac{1+i\eta}{1+(i+2)\eta}\;.
    \label{eqn:app_true_ratio}
\end{equation}
This ratio depends on $i$ unless $\eta=0$. Thus, exact perspective-projected ruler spacings are generally not a constant-ratio GP. This is why we avoid claiming that GP is an exact perspective invariant.

\noindent\textit{GP as a compact approximation.}
Although Eq.~\ref{eqn:app_true_projective_spacing} is not exactly a GP, it changes smoothly and monotonically over the visible ruler segment. RulerNet approximates this sequence using normalized GP spacings
\begin{equation}
    D_i^{\mathrm{GP}} = \rho^i\;,
    \label{eqn:app_gp_spacing}
\end{equation}
where the first interval has been normalized to one and a single scalar $\rho$ controls the compression or expansion of adjacent spacings. Equivalently, the unnormalized intervals can be written as $\Delta_i^{\mathrm{GP}}=\Delta_0\rho^i$. This is intentionally compact: after detecting and projecting marks onto a line, the scale model needs only an initial spacing and one ratio parameter to represent the dominant perspective trend. This low-dimensional structure is important for robust fitting because detected marks may be missing, duplicated, or slightly displaced.

For a transparent analytic error estimate, consider $N$ visible adjacent intervals and choose $\rho$ so that the GP matches the first and last true projected intervals. Since $D_0^{\mathrm{true}}=1$, this gives
\begin{equation}
    \rho
    =
    \left(D_{N-1}^{\mathrm{true}}\right)^{1/(N-1)}
    =
    \left[
    \frac{1+\eta}{(1+(N-1)\eta)(1+N\eta)}
    \right]^{\frac{1}{N-1}}\;.
    \label{eqn:app_rho_fit}
\end{equation}
The relative approximation error at interval $i$ is
\begin{align}
    \varepsilon_i
    &=
    \frac{D_i^{\mathrm{GP}}}{D_i^{\mathrm{true}}}-1 \nonumber\\
    &=
    \rho^i
    \frac{(1+i\eta)(1+(i+1)\eta)}{1+\eta}
    -1\;,
    \label{eqn:app_gp_error}
\end{align}
and the worst-case relative error over the visible ruler is
\begin{equation}
    E_{\max} = \max_{0\leq i < N} |\varepsilon_i|\;.
    \label{eqn:app_max_gp_error}
\end{equation}
This endpoint-matched GP is used here only to make the approximation error interpretable. In the actual RulerNet implementation, the GP parameters are fitted from detected image marks using Chamfer-distance optimization or predicted by DeepGP; both serve the same purpose of finding a compact constant-ratio approximation to the observed mark sequence.

\noindent\textit{Connection to camera geometry.}
Under a pinhole camera model, the denominator variation in Eq.~\ref{eqn:app_1d_homography} is governed by depth change along the ruler. Let $Z_0$ denote the depth of the first visible ruler mark from the camera, and let $\phi\in[0,90^\circ]$ denote the angle between the ruler direction and the image plane, where $\phi=0^\circ$ means the ruler is parallel to the image plane and the limiting case $\phi\to90^\circ$ means the ruler points along the viewing direction. Then the normalized perspective strength can be written as
\begin{equation}
    \eta = \frac{\ell\sin\phi}{Z_0}\;.
    \label{eqn:app_eta_angle}
\end{equation}
Therefore, GP approximation error decreases when the ruler is farther from the camera, when fewer intervals are visible, when centimeter spacing is small relative to depth, or when the ruler is closer to being parallel to the image plane.

Figure~\ref{fig:app_gp_perspective_error} plots $E_{\max}$ as a function of $\phi$ for $\ell=1$ cm, $N=10$ visible intervals, and representative close-range depths $Z_0\in\{25,30,40,50\}$ cm. Even in the conservative strong-perspective $\lim_{\phi\to90^\circ}\sin\phi=1$, the worst-case relative spacing errors are only 2.30\%, 1.68\%, 1.01\%, and 0.67\%, respectively. At $\phi=60^\circ$, the corresponding errors decrease to 1.79\%, 1.30\%, 0.77\%, and 0.51\%. These small errors support the use of GP as a practical approximation: it captures the dominant non-uniform scale trend induced by perspective while remaining compact enough to be fitted robustly from noisy ruler-mark detections and learned efficiently by a lightweight DeepGP model.
}

\begin{figure*}[!t]
    \centering
    \includegraphics[width=0.78\linewidth]{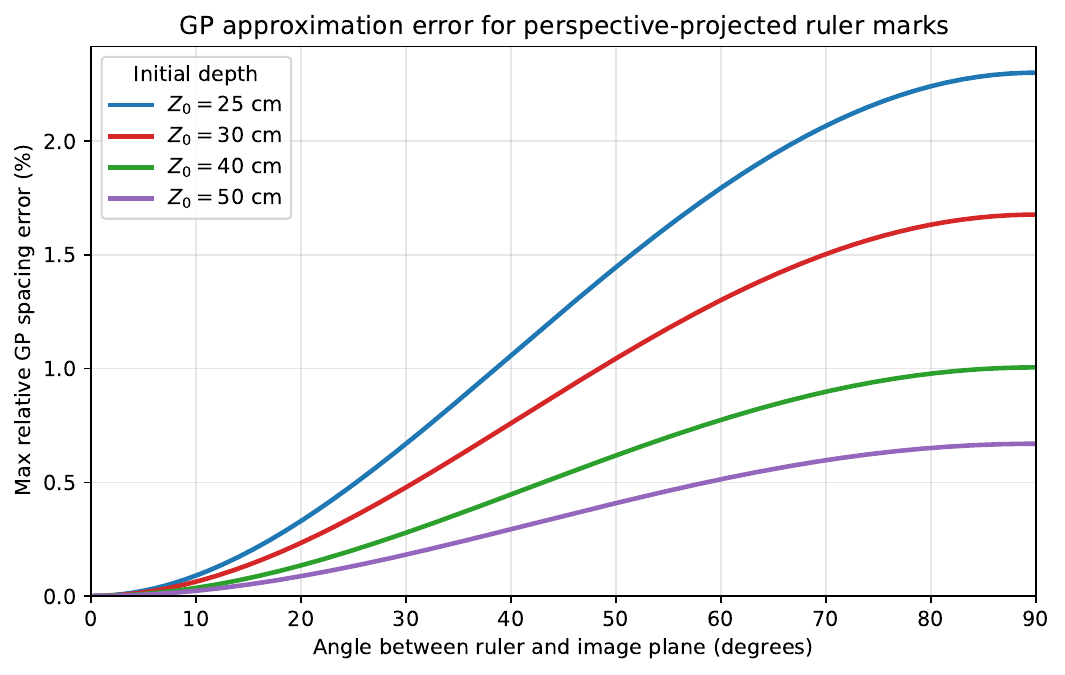}
    \caption{\revblue{Maximum relative error between the exact 1D projective ruler-spacing model and an endpoint-fitted GP approximation. The calculation uses 1 cm mark spacing and 10 visible adjacent intervals. Even under strong perspective, the GP approximation error remains small for typical close-range camera-ruler distances, supporting GP as a compact surrogate for projective spacing variation.}}
    \label{fig:app_gp_perspective_error}
\end{figure*}

\section{Dataset}
\begin{figure*}[!t]
    \centering
    \includegraphics[width=0.8\linewidth]{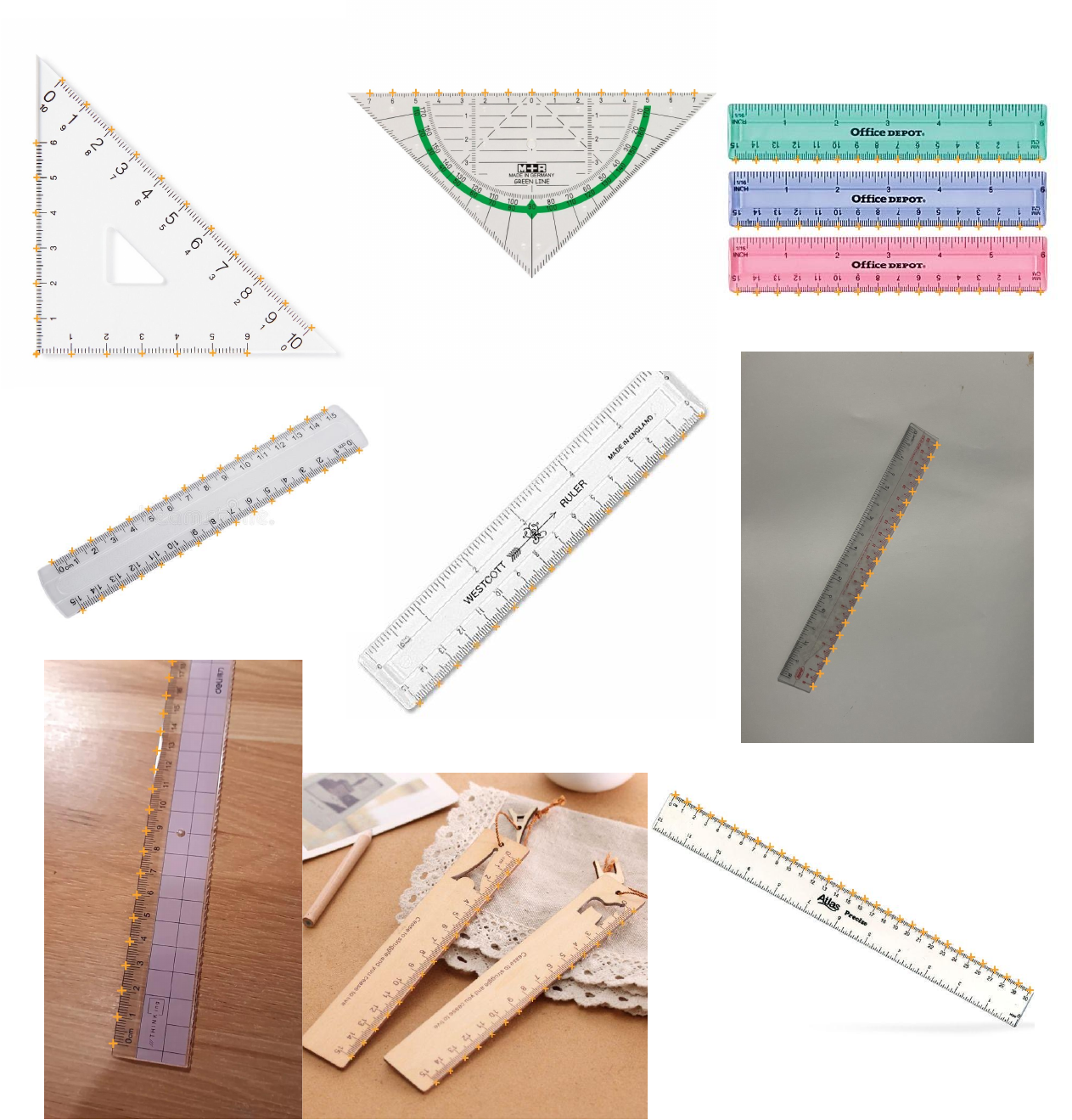}
    \caption{Examples from the AnyRuler dataset with our annotation.}
    \label{fig:app_anno_anyruler}
\end{figure*}

\begin{figure*}[!t]
    \centering
    \includegraphics[width=0.8\linewidth]{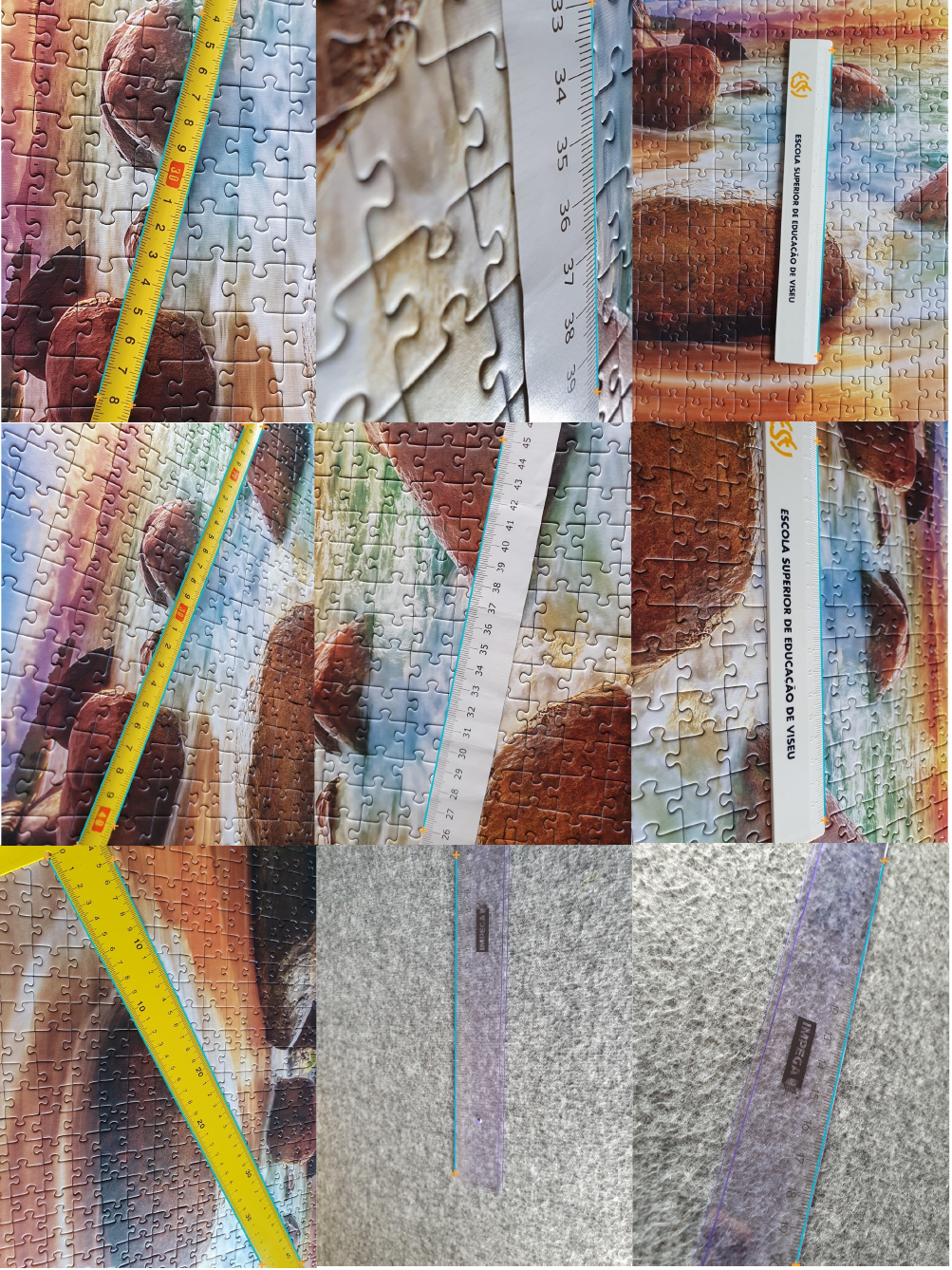}
    \caption{Examples from the Rulers2023~\citep{matuzevivcius2023rulers2023} dataset with our annotation.}
    \label{fig:app_anno_ruler2023}
\end{figure*}

We present example ruler images with our annotations from both the AnyRuler dataset (Fig.~\ref{fig:app_anno_anyruler}) and the Rulers2023 dataset~\citep{matuzevivcius2023rulers2023} (Fig.~\ref{fig:app_anno_ruler2023}). The AnyRuler dataset includes annotations indicating the positions of individual ruler marks, whereas the Rulers2023 dataset provides labeled line segments along with their corresponding lengths in centimeters.

\begin{figure*}
    \centering
    \includegraphics[width=0.8\linewidth]{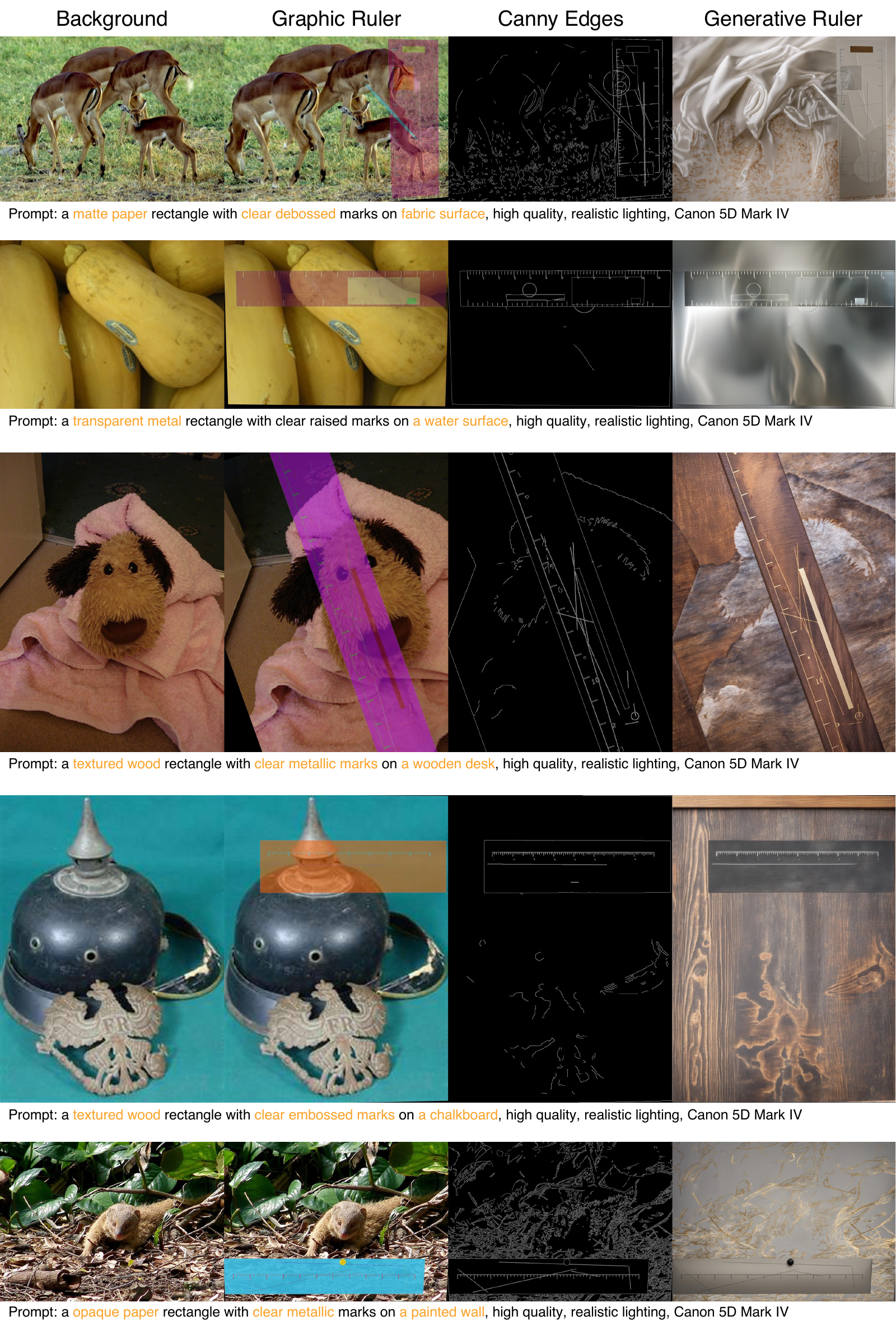}
    \caption{Example synthetic rulers.}
    \label{fig:app_synthetic_rulers}
\end{figure*}

We present additional synthetic rulers images generated by our pipeline in Fig.~\ref{fig:app_synthetic_rulers}. The positions of ruler marks are recorded since they are part of the parameterization of the graphics-based ruler generator. The prompts used for ControlNet are shown below each corresponding image.

\begin{figure*}
    \centering
    \includegraphics[width=0.7\linewidth]{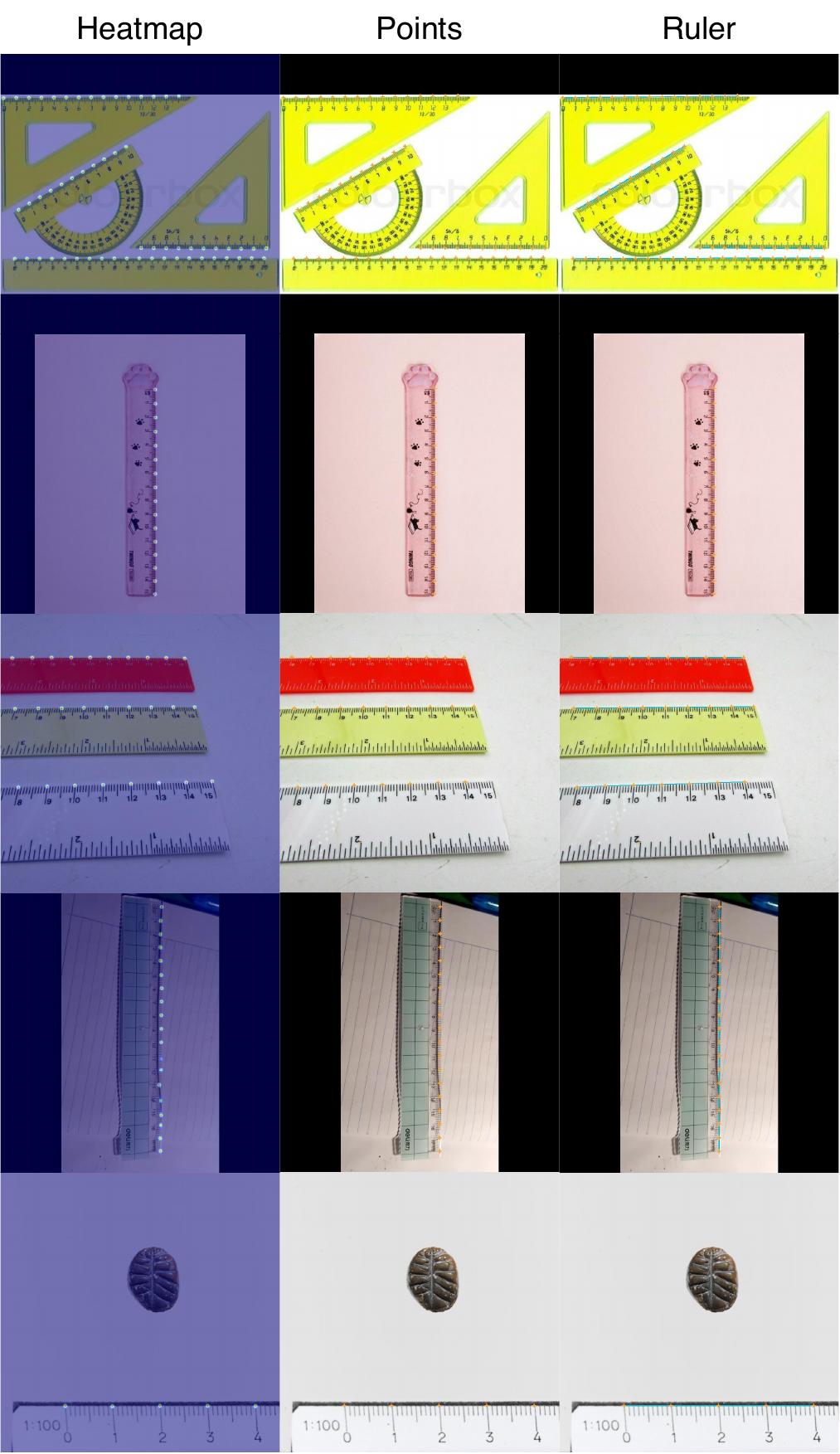}
    \caption{Additional qualitative results of RulerNet on AnyRuler dataset.}
    \label{fig:app_qual_anyruler}
\end{figure*}

\begin{figure*}
    \centering
    \includegraphics[width=0.7\linewidth]{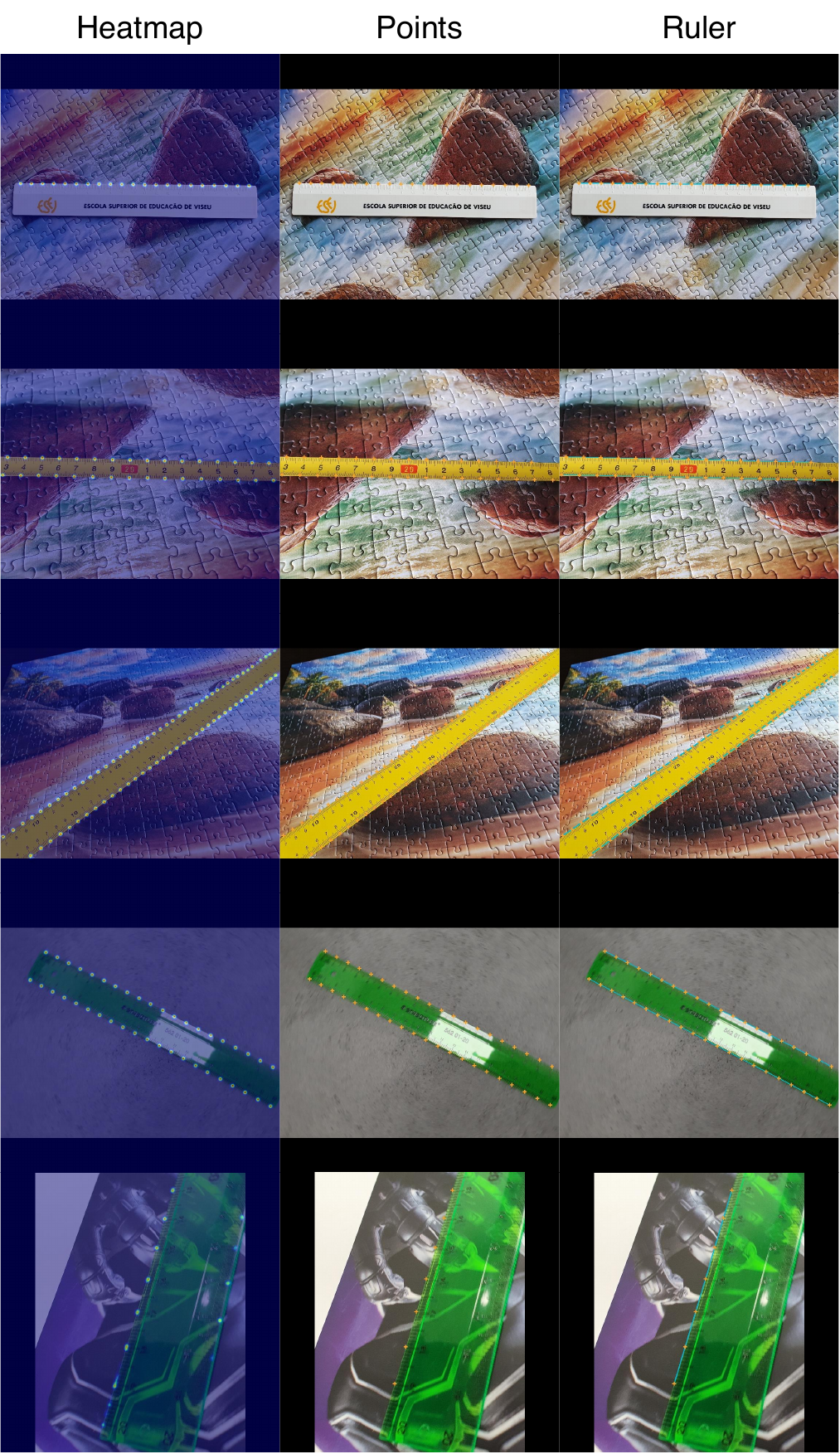}
    \caption{Additional qualitative results of RulerNet on Rulers2023~\citep{matuzevivcius2023rulers2023} dataset.}
    \label{fig:app_qual_ruler2023}
\end{figure*}

Furthermore, we present additional qualitative results of RulerNet on both the AnyRuler (Fig.~\ref{fig:app_qual_anyruler}) and Rulers2023 (Fig.~\ref{fig:app_qual_ruler2023}) datasets. For each example, the columns show: (1) the heatmap produced by the dense predictor, (2) the extracted ruler marks, and (3) the recovered rulers. 

\noindent\textit{Graphics-based Image Generator Parameters:} The parameters for the graphics-based ruler generator are listed in the Python-like function call in Listing~\ref{lst:draw_ruler}. We used a script to randomly generate these parameters when generating ruler images.

\noindent\textit{Available Values for the ControlNet Prompt Variables:} We used ChatGPT to generate a set of possible values for each variable in the ControlNet~\citep{zhang2023adding} prompt ``a \{visual\_appearance\} \{material\} rectangle with clear \{mark\_appearance\} marks on \{background\}.'' The available values for each variable are listed in Table~\ref{tab:app_prompt_variable} and were randomly sampled when generating ruler images.

\begin{table*}
\centering
\setlength{\tabcolsep}{2pt}
\caption{Available prompt variable values.}
\begin{tabular}{cccc}
\toprule
\textit{material} & \textit{visual\_appearance} & \textit{mark\_appearance} & \textit{background} \\ \midrule
plastic                & reflective   & engraved        & a wooden desk       \\
wood                   & transparent  & printed         & white paper         \\
metal                  & opaque       & embossed        & a blackboard        \\
glass                  & matte        & debossed        & metal surface       \\
rubber                 & glossy       & stamped         & a glass table       \\
composite materials    & textured     & laser-etched   & concrete floor      \\
paper                  & smooth       & painted         & a marble countertop \\
cardboard              & colored      & raised          & fabric surface      \\
                       & patterned    & indented        & a grass field       \\
                       & gradient     & dotted          & sandy surface       \\
                       & frosted      & striped         & a water surface     \\
                       & clear        & highlighted     & carpet              \\
                       &              & faded           & tile floor          \\
                       &              & bold            & a painted wall      \\
                       &              & thin            & a digital screen    \\
                       &              & dual-color      & a chalkboard        \\
                       &              & metallic        & cardboard sheet     \\
                       &              & contrasted      & a leather surface   \\
                       &              & glowing         & a plastic sheet     \\
                       &              & reflective      & the sky background  \\
                       &              &                 & a blurred background\\
                       &              &                 & none                \\ \bottomrule
\end{tabular}%
\label{tab:app_prompt_variable}
\end{table*}

\begin{listing*}[!ht]
\begin{minted}[linenos, breaklines, fontsize={\fontsize{8.5}{7.5}\selectfont}]{python}
draw_ruler(
    # Basic parameters
    image,  # The image canvas on which the ruler will be drawn.
    position,  # The starting (x, y) coordinates for drawing the ruler.
    length_cm,  # The total length of the ruler in centimeters.
    cm_to_px=37.7952755906,  # Conversion factor to convert centimeters to pixels.
    ruler_height_cm=3,  # Height of the ruler (in cm) as it appears on the image.
    ruler_extension_fraction=0.2,  # Fraction of the ruler length to extend beyond the ruler mark area.
    orientation="horizontal",  # Orientation of the ruler: "horizontal" or "vertical".

    # Style parameters
    fill_color=(255, 255, 255),  # RGB color tuple used to fill the ruler's background.
    edge_color=(0, 0, 0),         # RGB color tuple for drawing the ruler's edges.
    thickness=2,  # Thickness (in pixels) of the ruler's border/outline.
    alpha=0.5,  # Transparency level of the ruler (0 = fully transparent, 1 = fully opaque).

    # Mark interval parameters
    mm_mark_interval=1,  # Distance (in millimeters) between successive mm marks.
    inch_mark_interval="1/16",  # Interval between inch marks, expressed as a fraction of an inch.

    # Font parameters for cm and inch labels
    cm_font=cv2.FONT_HERSHEY_SIMPLEX,  # Font type for centimeter labels.
    inch_font=cv2.FONT_HERSHEY_SIMPLEX,  # Font type for inch labels.
    cm_font_scale=0.5,  # Scaling factor for the centimeter label font size.
    inch_font_scale=0.5,  # Scaling factor for the inch label font size.
    cm_font_color=(0, 0, 0),  # RGB color tuple for the centimeter label text.
    inch_font_color=(0, 0, 0),  # RGB color tuple for the inch label text.
    cm_font_offset_x=0,  # Horizontal offset for positioning centimeter labels.
    cm_font_offset_y=0,  # Vertical offset for positioning centimeter labels.
    inch_font_offset_x=0,  # Horizontal offset for positioning inch labels.
    inch_font_offset_y=0,  # Vertical offset for positioning inch labels.

    # Mark length parameters
    cm_mark_length=20,  # Length (in pixels) of the main centimeter marks.
    mm_mark_length=10,  # Length (in pixels) of the millimeter marks.
    half_cm_mark_length=15,  # Length (in pixels) of the half-centimeter marks.
    inch_mark_length=20,  # Length (in pixels) of the main inch marks.
    sub_inch_mark_length=10,  # Length (in pixels) for smaller divisions of an inch.
    half_inch_mark_length=15,  # Length (in pixels) of the half-inch marks.

    # Mark color parameters
    cm_mark_color=(50, 50, 50),  # RGB color tuple for centimeter marks.
    mm_mark_color=(50, 50, 50),  # RGB color tuple for millimeter marks.
    inch_mark_color=(100, 100, 100),  # RGB color tuple for inch marks.
    sub_inch_mark_color=(100, 100, 100),  # RGB color tuple for sub-inch marks.

    # Label display options
    show_cm_numbers=True,  # Boolean flag to display numeric labels for centimeters.
    show_inch_numbers=True,  # Boolean flag to display numeric labels for inches.

    # Perspective transform parameters
    tilt_factor_horizontal=0.1,  # Factor controlling horizontal tilt for a perspective effect.
    tilt_factor_vertical=0.1,  # Factor controlling vertical tilt for a perspective effect.

    # Random shape parameters
    num_random_lines=10,  # Number of random lines to add for decorative effects.
    num_random_shapes=10,  # Number of random shapes to add for decorative effects.
    
    # Additional side marks
    other_marks="none",  # Option to add extra side marks; choices: "none", "cm", "inch".
    mark_offset=0,  # Offset (in pixels) for positioning extra marks from the ruler edges.
    draw_cm_line=False,  # Boolean flag to draw a continuous line along the centimeter marks.
    cm_line_color=(0, 255, 0),  # RGB color tuple for the line drawn along centimeter marks.
    cm_line_thickness=2  # Thickness (in pixels) of the line drawn along centimeter marks.
)
\end{minted}
\caption{Python-like function to draw a ruler with various customization parameters.}
\label{lst:draw_ruler}
\end{listing*}

\section{Additional Results}\label{}
\revblue{We further tested different scale extraction methods on the final RulerNet model, which was pre-trained using all available data: the AnyRuler training data, 102,000 graphics-based samples drawn from the online programmatic generator, and 102,000 fixed pre-generated ControlNet-refined generative images. The model was then fine-tuned with the AnyRuler training data plus 1,000 synthetic samples per type, using online graphics-based generation and pre-generated generative images.} As shown in Table~\ref{tab:sup_abl_modules}, the GP method again achieved the best overall performance. Therefore, GP should be used regardless of the accuracy of the ruler mark localization module.

\begin{table}
    \centering
    \footnotesize
   \caption{Evaluation of RulerNet using different scale extraction methods, pre-trained and trained on the combination of synthetic data and AnyRuler dataset.}
    \begin{tabular}{lcc}
    \toprule
         &  AnyRuler & Rulers2023~\citep{matuzevivcius2023rulers2023}\\
         \midrule
        Direct & 1.50 & 1.82 \\
        Median & 1.33 & 2.08 \\
        GP & 1.37 & 1.50 \\
    \bottomrule
    \end{tabular}
    \label{tab:sup_abl_modules}
\end{table}

\subsection{\revblue{Inch-Ruler Adaptation}}
\revblue{The primary RulerNet model is trained and evaluated for metric rulers with centimeter marks. To test whether the framework can be adapted to another ruler unit when labeled data are available, we generated 1,000 synthetic inch-ruler images using the same SDXL ControlNet pipeline, converted them to the AnyRuler format with \texttt{target\_spacing\_in\_cm}=2.54, and used an 800/200 train/evaluation split. The detector was trained with the same CE+DICE objective for 200 epochs, and scale extraction used DeepGP. This experiment is an adaptation test, not a zero-shot evaluation of the released metric-ruler model.}

\begin{table}[!t]
    \centering
    \footnotesize
    \caption{\revblue{Synthetic inch-ruler adaptation results on a 200-image evaluation split. Pixel/cm errors account for the 2.54 cm spacing between adjacent whole-inch marks.}}
    \begin{tabular}{lc}
    \toprule
    \revblue{Metric} & \revblue{Value} \\
    \midrule
    \revblue{Mean pixel/cm error} & \revblue{0.242470} \\
    \revblue{Median per-sample pixel/cm error} & \revblue{0.065415} \\
    \revblue{95th percentile pixel/cm error} & \revblue{0.771623} \\
    \revblue{Chamfer distance} & \revblue{3.197235} \\
    \revblue{GT marks per image} & \revblue{7--16, median 10} \\
    \revblue{NaN scale predictions} & \revblue{0} \\
    \bottomrule
    \end{tabular}
    \label{tab:app_inch_adaptation}
\end{table}

\section{Analysis of Failure Cases}
\begin{figure*}
    \centering
    \includegraphics[width=0.95\linewidth]{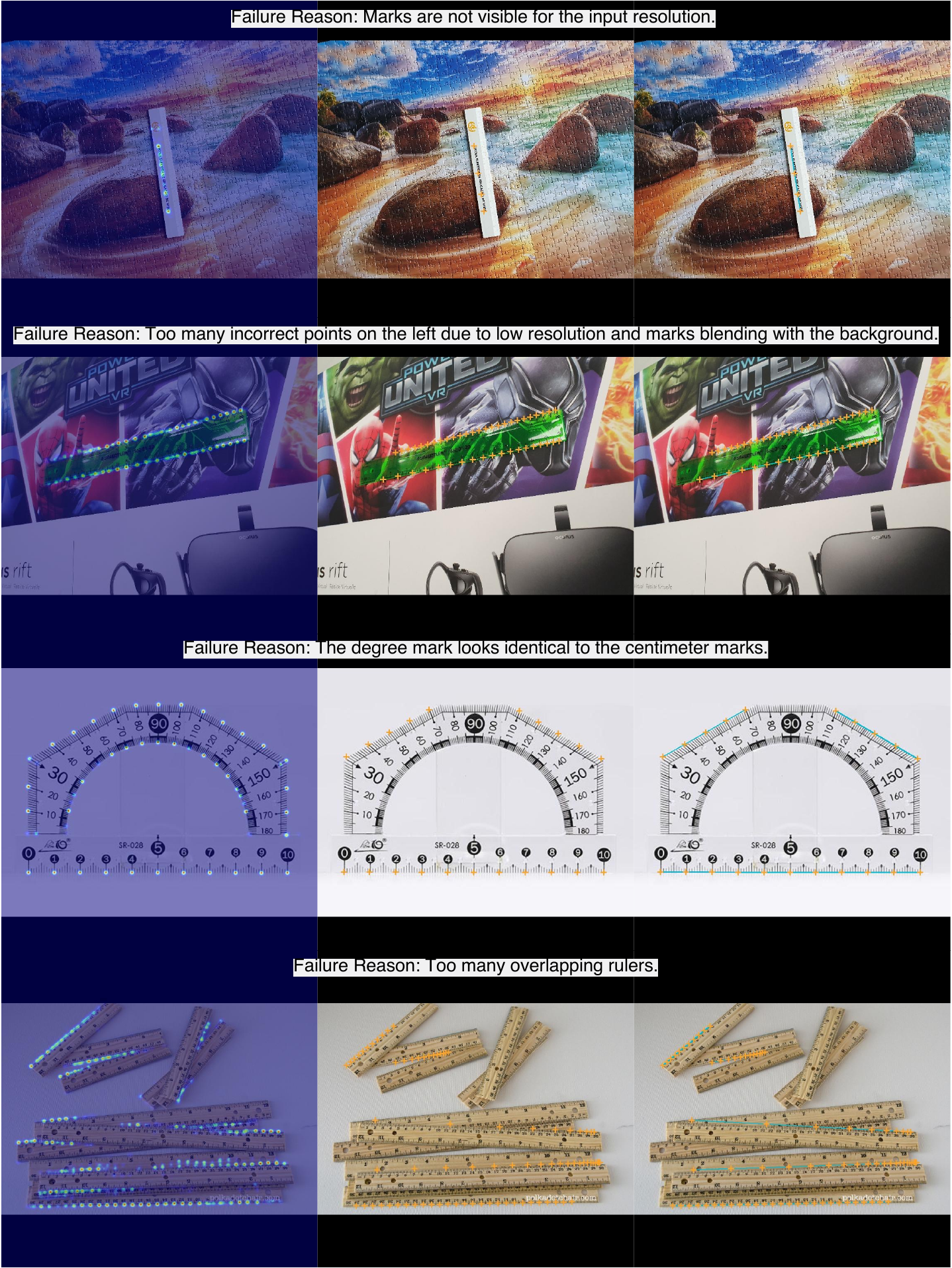}
    \caption{Examples of failure cases from both the AnyRuler and Rulers2023~\citep{matuzevivcius2023rulers2023} datasets.}
    \label{fig:app_failure_cases}
\end{figure*}

Despite the strong robustness of RulerNet, there are still scenarios in which the model can become confused. We visualize several representative failure cases in Fig.~\ref{fig:app_failure_cases}. For instance, when ruler marks are not clearly visible (Row 1), RulerNet understandably fails to localize them. Transparent rulers placed against complex or cluttered backgrounds can also reduce mark visibility, as blending effects make it difficult for the model to distinguish the marks from the surroundings (Row 2). Additionally, objects or patterns that visually resemble centimeter markings (degree marks) may produce false positives (Row 3). In crowded scenes with multiple overlapping rulers, the model may struggle to disambiguate which marks belong to which ruler (Row 4).

To ensure accurate predictions in practice, users should consider the following recommendations: 
\begin{itemize} 
\item Use rulers with high-contrast markings that are clearly visible. 
\item Avoid transparent or reflective rulers in visually complex environments. 
\item Minimize background clutter near the ruler, especially textures or objects that resemble measurement marks. 
\item Ensure that the ruler lies flat and is not significantly occluded or distorted. 
\item If multiple rulers are present in a single frame, avoid overlapping when possible. \end{itemize}

By following these guidelines, users can significantly reduce the likelihood of failure and help RulerNet achieve the most reliable scale estimation in real-world settings.

\end{document}